\documentclass[graybox]{svmult}

\usepackage{mathptmx}       
\usepackage{helvet}         
\usepackage{courier}        
\usepackage{type1cm}        
\usepackage{makeidx}         
\usepackage{graphicx}        
\usepackage{multicol}        
\usepackage[bottom]{footmisc}

\makeindex             

\usepackage{graphicx}
\usepackage{amsmath,amssymb,amsfonts}

\newtheorem{prop}{Proposition}

\usepackage{listings}
\newcommand{\ie}{i.e.~}

\newcommand{\eg}{e.g.~}
\newcommand{\al}{al.~}
\newcommand{\espace}{\vspace{10pt}}
\newcommand{\cpsm}{CPSM}
\newcommand{\cpsmemb}{CPSM-emb}
\newcommand\eqdef{\mathrel{\stackrel{\makebox[0pt]{\mbox{\normalfont\tiny def}}}{=}}}

\newcommand{\sysfont}{\textit}
\newcommand{\asprin}{\sysfont{asprin}}
\newcommand{\clasp}{\sysfont{clasp}}
\newcommand{\clingo}{\sysfont{clingo}}

\newcommand{\dlv}{\sysfont{dlv}}
\newcommand{\gecode}{\sysfont{gecode}}

\newcommand{\smodels}{\sysfont{smodels}}

\usepackage{breakcites}

\begin{document}

\title*{Efficiency Analysis of ASP Encodings for Sequential Pattern Mining Tasks}
\author{Thomas Guyet and Yves Moinard and Ren\'e Quiniou and Torsten Schaub}
\institute{Thomas Guyet \at AGROCAMPUS-OUEST/IRISA-UMR 6074/Inria, \email{thomas.guyet@agrocampus-ouest.fr}
\and Yves Moinard \at Inria -- Centre de Rennes \email{yves.moinard@inria.fr}
\and Ren\'e Quiniou \at Inria -- Centre de Rennes \email{rene.quiniou@inria.fr}
\and Torsten Schaub \at Potsdam University/Inria \email{torsten@cs.uni-potsdam.de}
}
%
%
\maketitle

\abstract{This article presents the use of Answer Set Programming (ASP) to mine sequential patterns. 
ASP is a high-level declarative logic programming paradigm for high level encoding combinatorial and optimization problem solving as well as knowledge representation and reasoning. Thus, ASP is a good candidate for implementing pattern mining with background knowledge, which has been a data mining issue for a long time. 
We propose encodings of the classical sequential pattern mining tasks within two  representations of embeddings (\emph{fill-gaps} vs \emph{skip-gaps}) and for various kinds of patterns: frequent, constrained and condensed. 
We compare the computational performance of these encodings with each other to get a good insight into the efficiency of ASP encodings. The results show that the \emph{fill-gaps} strategy is better on real problems due to lower memory consumption. Finally, compared to a constraint programming approach (\cpsm), another declarative programming paradigm, our proposal showed comparable performance.}

\abstract*{This article presents the use of Answer Set Programming (ASP) to mine sequential patterns. 
ASP is a high-level declarative logic programming paradigm for high level encoding combinatorial and optimization problem solving as well as knowledge representation and reasoning. Thus, ASP is a good candidate for implementing pattern mining with background knowledge, which has been a data mining issue for a long time. 
We propose encodings of the classical sequential pattern mining tasks within two representations of embeddings (\emph{fill-gaps} vs \emph{skip-gaps}) and for various kinds of patterns: frequent, constrained and condensed. 
We compare the computational performance of these encodings with each other to get a good insight into the efficiency of ASP encodings. The results show that the \emph{fill-gaps} strategy is better on real problems due to lower memory consumption. Finally, compared to a constraint programming approach (\cpsm), another declarative programming paradigm, our proposal showed comparable performance.}

\section{Introduction} \label{sec:intro}
Pattern mining is a data analysis task aiming at identifying ``meaningful'' patterns in a database of structured data (\eg itemsets, sequences, graphs).
Sequential pattern mining consists in discovering subsequences as patterns in a sequence database \cite{Shen2014}. This problem has been introduced at the early beginning of the pattern mining field \cite{Agrawal1995} with the itemsets mining task \cite{Agrawal1993}. 
Sequential pattern mining is known to have a higher complexity than itemsets mining, but it has broad applications \cite{SPMapplication}. It includes -- but is not limited to -- the analysis of patient care pathways, education traces, digital logs (web access for client profiling, intrusion detection from network logs), customer purchase (rules for purchases recommendations), text and bioinformatic sequences.

In most cases, ``interesting" patterns are the frequent ones. A pattern is said to be frequent if it appears at least $f_{min}$ times in the database, where $f_{min}$ is a frequency threshold given by the data analyst.
This interestingness criterion reveals some important behaviours in the datasets and, above all, it benefits from an interesting property (anti-monotonicity) that make algorithms very efficient, even on large databases.
Two decades of research on the specific task of frequent sequential pattern mining have led to very efficient mining methods and strategies to extract the complete set of frequent patterns or condensed representation of frequent patterns \cite{BIDE2004}. These methods can currently process millions of sequences with very low frequency threshold.

The challenge of mining a deluge of data is about to be solved, but is also about to be replaced by another issue: the deluge of patterns.
In fact, the size of the complete set of frequent patterns explodes with the database size and density \cite{Lhote2010}.
The data analyst cannot handle such volumes of results. A broad range of research, from data visualization \cite{perer2014frequence} to database sampling \cite{low2013mining}  is currently attempting to tackle this issue. 
The data-mining community has mainly focused on the addition of expert constraints on sequential patterns \cite{Pei2004}.

Recent approaches have renewed the field of Inductive Logic Programming \cite{muggleton1994inductive} by exploring declarative pattern mining. 
Similarly, some works have tackled the itemset mining task \cite{guns2015miningzinc,jarvisalo2011itemset}. Recently, some propositions have extended the approach to sequence mining \cite{Negrevergne15, coquery2012sat, Metivier2013}. Their practical use depends on the efficiency of their encoding to process real datasets. 
Thanks the improvements on satisfiability (SAT) or constraints programming (CP) solving techniques and solvers, such approaches become realistic alternatives for highly constrained mining tasks. Their computational performances closely reach those of dedicated algorithms.

\espace

The long term objective is to benefit from the genericity of solvers to let a user specify a potentially infinite range of constraints on the patterns. Thus, we expect to go from specific algorithm constraints to a rich query language for pattern mining. 

\espace

The approach we propose in this paper uses the formalism of Answer Set Programming (ASP) and the solver \clingo. ASP is a logic programming language, as Prolog. Its first order syntax makes ASP programs easy to understand. Furthermore, ASP benefits from efficient solvers to compute efficiently the solution answer sets \cite{lifschitz:aaai:2008}. 

The contributions of this article are twofold. 1) The article presents a declarative approach which provides a high-level specification of a broad range of sequential pattern mining tasks in a unique framework. It demonstrates that this mining task and its related problems -- mining closed, maximal and constrain patterns -- can easily be encoded with pure declarative ASP. 2) The article extensively evaluates the proposed encodings to draw the computational strengths and limits of ASP for declarative pattern mining.
It gives also experimental results about time/memory computing efficiency of the solving process and provides alternative encodings to improve the computing efficiency. The proposed encodings are compared to the results of the \cpsm\ software, based on CP programming \cite{Negrevergne15}.

\espace

The article is organized as follows: Sect. \ref{sec:ASP} introduces ASP programming, its principles and the solver \clingo. Then in Sect. \ref{sec:spm}, we introduce sequential pattern mining. 
In Sect. \ref{sec:spm_with_asp}, we give several ASP encodings of the basic sequential pattern mining task.
Sect. \ref{sec:alternative_spm_asp} presents encodings for alternative sequential pattern mining tasks, including the use of constraints and the extraction of condensed representations.
After presenting some related works in Sect. \ref{sec:soa}, we present our experiments in Sect. \ref{sec:experiments}.

\section{ASP -- Answer Set Programming}\label{sec:ASP}
In this section we introduce the Answer Set Programming (ASP) paradigm, syntax and tools. Sect. \ref{sec:aspprinc} introduces the main principles and notations of ASP. Sect. \ref{sec:aspex} illustrates them on the well-known graph coloring problem.

\subsection{Principles of Answer Set Programming}\label{sec:aspprinc}
ASP is a declarative programming paradigm. 
From a general point of view, declarative programming gives a description of what is a problem instead of specifying how to solve it.
Several declarative paradigms have been proposed, differing in the modelling formalism they use. For instance, logic programming \cite{lallouet:hal-00758896} specifies the problem using a logic formalism, the SAT para\-digm encodes the problem with boolean expressions \cite{biere2009handbook}, the CP (constraint programming) paradigm specifies the problem using constraints \cite{rossi2006handbook}.
ASP belongs to the class of logic programming paradigms, such as Prolog. 
The high-level syntax of logic formalisms makes generally the program easier to understand than other declarative programming paradigms.

\espace

An \emph{ASP program} is a set of rules of the form
\begin{align}\label{eq:rule}\tt
a_0 \mathrel{\texttt{:-}} a_1,\ldots,a_m, \text{\lstinline!not! } a_{m+1},\ldots,\text{\lstinline!not! } a_n.
\end{align}

where each $\mathtt{a_i}$ is a propositional atom for $\mathtt{0}\leq\mathtt{i}\leq\mathtt{n}$ and \lstinline!not! stands for \emph{default negation}. In the body of the rule, commas denote conjunctions between atoms.
Contrary to Prolog, the order of atoms is meaningless.
In ASP, rule \eqref{eq:rule}  may be interpreted as ``\emph{if $\mathtt{a_1,\ldots,a_m}$ are all true and if none of $\mathtt{a_{n+1},\ldots,a_n}$ can be proved to be true, then $\mathtt{a_0}$ is true.}"

If $\mathtt{n}=\mathtt{0}$, \ie the rule body is empty,  \eqref{eq:rule} is called a \emph{fact} and the symbol ``\texttt{:-}'' may be omitted. Such a rule states that the atom $a_0$ has to be true.
If $\mathtt{a_0}$ is omitted, \ie the rule head is empty,  \eqref{eq:rule} represents an integrity constraint. 

Semantically, a logic program induces a collection of so-called \emph{answer sets},
which are distinguished models of the program determined by answer sets semantics;
see \cite{gellif91a} for details.
For short, a model assigns a truth value to each propositional atoms of the program and this set of assignments is valid. An answer set is a minimal set of true propositional atoms that satisfies all the program rules.
Answer sets are said to be minimal in the way that only atoms that have to be true are actually true.

\espace

To facilitate the use of ASP in practice, several extensions have been developed. 
First of all, rules with \emph{variables} are viewed as shorthands for the set of their ground instances.
This allows for writing logic programs using a first order syntax. Such kind of syntax makes program shorter, but it hides the grounding step and its specific encoding issues, especially from the memory management point of view.

Further language constructs include \emph{conditional literals} and \emph{cardinality constraints} \cite{siniso02a}.
The former are of the form
\[\tt
a\mathrel{:}{b_1,\dots,b_m}
\]
the latter can be written as
\[\tt
s~\{c_1;\dots;c_n\}~t
\]
where $\mathtt{a}$ and $\mathtt{b_i}$ are possibly default negated literals for $0\leq i\leq m$,
and each $\mathtt{c_j}$ is a conditional literal for $1\leq i\leq n$. The purpose of conditional literals is to govern the instantiation of a literal \texttt{a} through the literals $\tt{b_1,\dots,b_m}$. In a cardinality constraint,
$\mathtt{s}$ (resp. $\mathtt{t}$) provides the lower (resp. upper) bound on the number of literals from $\tt{c_1;\dots;c_n}$ that must be satisfied in the model.

A cardinality constraint in the head of the rule defines a specific rule called a \textit{choice rule}:
\[\tt
s~\{c_1;\dots;c_n\}~t \mathrel{\texttt{:-}} a.
\]
If $\mathtt{a}$ is true then all atoms of a subset $\mathcal{S} \subset \{c_1,\dots,c_n\}$ of size between $s$ and $t$ have to be true. All such subsets are admissible according to this unique rule, but not in the same model. All such subsets contribute to alternative answer sets.
It should be noted that alternative models are solved independently. It is not possible to specify constraints that involve several models.

ASP problem solving is ensured by efficient solvers \cite{lifschitz:aaai:2008} which are based on the same technologies as constraint programming solvers or satisfiability checking (SAT) solvers. 
\smodels\ \cite{syrjanen2001smodels}, \dlv\ \cite{Leone2006}, \sysfont{ASPeRiX} \cite{Asperix} or \clingo\ \cite{gekakaosscsc11a} are  well-known ASP solvers. Due to the computational efficiency it has demonstrated and its broad application to real problems, we use \clingo\ as a basic tool for designing our encodings.

\espace

The basic method for  programming in ASP is to follow a \textit{generate-and-test} methodology.
Choice rules generate solution candidates, while integrity constraints are tested to eliminate those candidates that violate the constraints.
The programmer should not have any concern about how solutions are generated. He/she just has to know that all possible solutions will be actually evaluated.
From this point of view, the ASP programming principle is closer to CP programming than to Prolog programming.
Similarly to these declarative programming approaches, the  difficulty of programming in ASP lies in the choices for the best way to encode problem constraints: it must be seen as the definition of the search space (\emph{generate} part) or as an additional constraint on solutions within this search space (\emph{test} part). This choices may have a large impact on the efficiency of the problem encoding.

\subsection{A simple example of ASP program}\label{sec:aspex}
\begin{figure}[tp]
\centering
\includegraphics[width=0.5\linewidth]{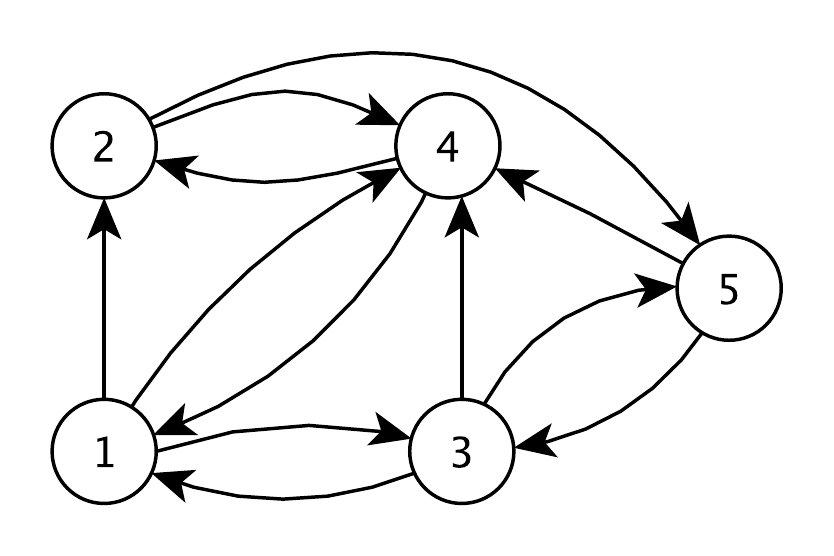}
\caption{An example graph for the graph coloring problem.}
\label{fig:graph}
\end{figure}

The following example illustrates the ASP syntax on encoding the graph coloring problem.
Lines 9-10 specify the problem as general rules that solutions must satisfy while lines 1-6 give the input data that defines the problem instance related to the graph in Fig. \ref{fig:graph}.

The problem instance is a set of colors, encoded with predicates \lstinline!col/1! and a graph, encoded with predicates \lstinline!vertex/1! and \lstinline!edge/2!.
The input graph has 5 vertices numbered from 1 to 5 and 12 edges. The 5 fact-rules describing the vertex are listed in the same line (Line 2).
It should be noted that, generally speaking, \lstinline!edge(1,2)! is different from \lstinline!edge(2,1)!, but considering that integrity constraints for the graph coloring problem are symmetric, it is sufficient to encode directed edge in only one direction.
Line 6 encodes the 3 colors that can be used: \lstinline!r!, \lstinline!g! and \lstinline!b!. Lower case letters represent values, internally encoded as integers, while strings beginning with upper case letters represent variables (see line 9 for instance).

\begin{lstlisting}[caption={Encoding of graph coloring -- ASP syntax and encoding example}, label=list:graph_coloring, float]
% instance of the problem: the graph and colors
vertex(1). vertex(2). vertex(3). vertex(4). vertex(5).
edge(1,2). edge(1,3). edge(1,4). edge(2,4). edge(2,5).
edge(3,1). edge(3,4). edge(3,5). edge(4,1). edge(4,2).
edge(5,3). edge(5,4).
col(r). col(b). col(g).

% graph coloring problem specification
1 { color(X, C) : col(C) } 1 :- vertex(X).
:- edge(X, Y), color(X, C), color(Y, C).
\end{lstlisting}

Lines 9 and 10 specify the graph coloring problem. The predicate \lstinline!color/2! encodes the color of a vertex:  \lstinline!color(X,C)! expresses that vertex \lstinline!X! has color \lstinline!C!.
Line 10 is an integrity constraint. It forbids neighbor vertices \lstinline!X! and \lstinline!Y! to have the same color \lstinline!C!\footnote{It is important to notice that the scope of a variable is the rule and each occurrence of a variable in a rule represents the same value.}.
The ease of expressing such integrity constraints is a major feature of ASP.

Line 9 is a choice rule indicating that for a given vertex \lstinline!X!, an answer set must contain exactly one atom of the form \lstinline!color(X,C)! where \lstinline!C! is a color. 
The grounded version of this rule is the following:
\begin{lstlisting}[numbers=none]
1 { color(1, r), color(1, b), color(1, g) } 1.
1 { color(2, r), color(2, b), color(2, g) } 1.
1 { color(3, r), color(3, b), color(3, g) } 1.
1 { color(4, r), color(4, b), color(4, g) } 1.
1 { color(5, r), color(5, b), color(5, g) } 1.
\end{lstlisting}

The variable \lstinline!X! is expanded according to the facts in line 2 and for each vertex, a specific choice rule is defined. Within the brackets, the variable \lstinline!C! is expanded according to the conditional expression in the rule head of line 9: the only admissible values for \lstinline!C! are color values.
For each line of the grounded version of the program, one and only one atom within brackets can be chosen. This corresponds to a unique mapping of a color to a vertex.
Line 9 can be seen as a search space generator for the graph coloring problem.

The set \lstinline!color(1,b) color(2,r) color(3,r) color(4,g) color(5,b)! is an answer set for the above program (among several others).

\espace

For more detailed presentation of ASP programming paradigm, we refer the reader to recent article of Janhunen and Nimeläthe \cite{Janhunen16}.

\subsection{The \sysfont{Potassco} collection of ASP tools}
The \sysfont{Potassco} collection is a set of tools for ASP developed at the University of Potsdam. The main tool of the collection is the ASP solver \clingo\  \cite{gekakaosscsc11a}. This solver offers both a rich syntax to facilitate encodings\footnote{\clingo\ is fully compliant with the recent ASP standard:\\
\url{https://www.mat.unical.it/aspcomp2013/ASPStandardization}} and a good solving efficiency. It is worth-noting that the ASP system \clingo\ introduced many facilities to accelerate the encoding of ASP programs.
For the sake of simplicity, we do not use them in the presented programs. A complete description of the \clingo\ syntax can be found in \cite{gekakasc14b}.

The \clingo\ solving process follows two consecutive main steps:
\begin{enumerate}
\item  \emph{grounding} transforms the initial ASP program into a set of propositional clauses, cardinality constraints and optimisation clauses. Note that grounding is not simply a systematic problem transformation. It also simplifies the rules to generate the as short as possible equivalent grounded program.
\item \emph{solving} consists in finding from one to all solutions of the grounded program. This step is performed by \clasp\  which is a conflict-driven ASP solver. The primary \clasp\ algorithm relies on conflict-driven \emph{nogood} learning. 
It is further optimized using sophisticated reasoning and implementation techniques, some specific to ASP, others borrowed from CDCL-based SAT solvers.
\end{enumerate}

The overall process may be controlled using procedural languages, \eg \emph{Python} or \emph{lua} \cite{gekakasc14b}. These facilities are very useful to automate processes and to collect statistics on solved problems.
Despite this procedural control which enables to interact with the grounder or the solver, it is important to note that once a program has been grounded, it cannot be changed.

\section{Sequential pattern mining: definition and notations}\label{sec:spm}
Briefly, the sequential pattern mining problem consists in retrieving from a sequence database $\mathcal{D}$ every frequent non empty sequence $P$, so-called a sequential pattern. Sequences, either in $\mathcal{D}$ or sequential patterns $P$, are multiset sequences of itemsets over a fixed alphabet of symbols (also called items). A pattern is frequent if it is a subsequence of at least $f_{min}$ sequences of $\mathcal{D}$, where $f_min$ is an arbitrary given threshold.
In this section, we introduce classical definitions and notations for frequent sequential pattern mining which will be useful to formulate the problem in an ASP setting. 
In the sequel, if not specified otherwise, a pattern is a sequential pattern.

\subsection{Sequences}\label{seq:sequences}
We introduce here the basic definitions of sequences of itemsets.
 $[n]=\{1, \dots, n\}$ denotes the set of the first $n$ strictly positive integers.

Let $(\mathcal{I}, =, <)$ be the set of items (alphabet) with a total order (\eg lexicographic order). 
An \emph{itemset} $A=(a_1, a_2, ..., a_n)$ is a subset of distinct increasingly ordered items from $\mathcal{I}$:  $\forall i \in [n],\; a_i\in\mathcal{E}\wedge \forall i \in [n-1],\; a_i<a_{i+1} \wedge \forall i,j \in [n],\; i \neq j \implies a_i\neq a_j$.
An itemset $\beta=(b_i)_{i\in [m]}$ is a sub-itemset of $\alpha=(a_i)_{i\in [n]}$, denoted $\beta \sqsubseteq \alpha$, iff there exists a sequence of $m$ integers $1 \leq i_1 < i_2 < ...  < i_m \leq n$ such that $\forall k \in [m], b_k=a_{i_k}$.
A \emph{sequence} $S$ is an ordered set of itemsets $S = \langle s_1, s_2, ..., s_n\rangle$: $\forall i,j \in [n],\; i < j$ means that $s_i$ occurs before $s_j$. The length of sequence $S$, denoted $|S|$, is equal to its number of itemsets. Two sequences $S= \langle s_1, s_2, ..., s_n\rangle$ and $T= \langle t_1, t_2, ..., t_m\rangle$ are equal iff $n=m$ and $\forall i \in [n],\; s_i=t_i$.

$T= \langle t_1, t_2, ..., t_m\rangle$ is a \emph{sub-sequence} of $S= \langle s_1, s_2, ..., s_n\rangle$, denoted $T \preceq S$, iff there exists a sequence of integers $1 \leq i_1 < i_2 < ...  < i_m \leq n$ such that $\forall k \in [m],\; t_k \sqsubseteq s_{i_k}$. In other words, $(i_k)_{1\leq k \leq m}$ defines a mapping from $[m]$, the set of indexes of $T$, to $[n]$, the set of indexes of $S$.
 We denote by $T \prec S$ the strict sub-sequence relation such that $T \preceq S$ and $T \neq S$.

$T=\langle t_1, t_2, ..., t_m\rangle$ is a  \emph{prefix} of $S= \langle s_1, s_2, ..., s_n\rangle$, denoted $T \preceq_b S$, iff $\forall i \in [k-1],\; t_i=s_{i}$ and $t_m \sqsubseteq s_{m}$. Thus, we have $T \preceq_b S \implies T \preceq S$.

A sequence $T$ \emph{is supported} by a sequence $S$ if $T$ is a sub-sequence of $S$, \ie $T \preceq S$.

\begin{example}[Sequences, subsequences and prefixes]
Let $\mathcal{I}=\{a,b,c\}$ with a lexicographic order ($a<b$, $b<c$) and the sequence $S=\langle a(bc)(abc)cb\rangle$. To simplify the notation, parentheses are omitted around itemsets containing only one item. The length of $S$ is 5. $\langle abb\rangle$, $\langle (bc)(ac)\rangle$ or $\langle a(bc)(abc)cb\rangle$ are sub-sequences of $S$. $a$, $\langle a(bc)\rangle$ and $\langle a(bc)a\rangle$ are prefixes of $S$.
\end{example}

\begin{prop}{}\label{prop:relationordre}
$\prec$ and $\prec_b$ induces two partial orders on the sequence set. For all sequences $(s,s')$, $s\prec_b s' \implies s\prec s'$.
\end{prop}

\subsection{Sequential pattern mining}

Let $\mathcal{D}= \lbrace S_1, S_2, \ldots , S_N \rbrace$ be a set of sequences. $\mathcal{D}$ is called a \emph{sequence database}.
The \emph{support of a sequence} $S$ in $\mathcal{D}$, denoted by $supp_{\mathcal{D}}(S)$, is the number of sequences of $\mathcal{D}$ that support $S$:
$$supp_{\mathcal{D}}(S)= |\{S_i \in \mathcal{D} | S \prec S_i \}|$$

\begin{prop}{}
$supp_{\mathcal{D}}(.)$ is an anti-monotonic measure on the set of subsequences of a sequence database $\mathcal{D}$ structured by $\prec$ or $\prec_b$.
\end{prop}
This proposition implies that for all pairs of sequences $P$, $Q$: $$P \prec Q \implies supp_{\mathcal{D}}(P) \geq supp_{\mathcal{D}}(Q)\text{, and}$$
$$P \prec_b Q \implies supp_{\mathcal{D}}(P) \geq supp_{\mathcal{D}}(Q).$$

Let $f_{min}$ be a \emph{frequency threshold} defined by the analyst. For any sequence $S$, if $supp_{\mathcal{D}}(S) \geq f_{min}$, we say that $S$ is a \emph{frequent sequence} or a \emph{(frequent) sequential pattern} of $\mathcal{D}$.
\emph{Mining sequential patterns} consists in extracting all frequent subsequences in a sequence database $\mathcal{D}$.

Every pattern mining algorithm \cite{Agrawal1995,BIDE2004,pei2007seqmining} uses the anti-monotonicity property to browse efficiently the pattern search space. In fact, this property ensures that a sequence $P$ including a sequence $Q$ which is not frequent, cannot be frequent itself. So, the main idea of classical algorithms is to extend the patterns until identifying a non frequent pattern.

\begin{example}[Sequential pattern mining]\label{ex:seq_mining}
To illustrate the concepts introduced above, we consider the following sequence database $\mathcal{D}$ containing sequences built on items in $\mathcal{I}=\{a,b,c,d\}$ such that $a< b$, $b< c$ and $a< c$. In this running example, and in the rest of this article, we focus on \emph{simple} sequences of items instead of sequences of itemsets.

\begin{center}
\begin{tabular}{cc}
\hline
SeqId & Sequence \\ \hline
1 & $\langle a c \rangle$ \\
2 & $\langle d a b c \rangle$ \\
3 & $\langle b \rangle$ \\
4 & $\langle a b c \rangle$ \\
5 & $\langle a b \rangle$ \\
6 & $\langle a c b c \rangle$ \\
7 & $\langle a b c \rangle$ \\ \hline
\end{tabular}
\end{center}

Given a threshold value $f_{min} = 3$, the frequent sequential patterns are: $\langle a \rangle$, $\langle b \rangle$, $\langle c \rangle$, $\langle ac \rangle$, $\langle bc \rangle$, $\langle ab \rangle$ and $\langle abc \rangle$.
\end{example}

It is interesting to relate the sequential pattern mining task with the presentation of ASP principles. The sequential pattern mining task rises two issues: 1) exploring a large search space, \ie the potentially infinite set of sequences and 2) assessing the frequency constraint (with respect to the given database).
Thus, sequential pattern mining can be considered as a \emph{generate and test} process which makes it straightforward to encode the mining task using ASP principles: 1) choice rules will define the search space and 2) the frequency assessment will be encoded using integrity constraints.

\section{Frequent sequential pattern mining with ASP}\label{sec:spm_with_asp}
In this section, we present several ASP encodings for sequential pattern mining. We assume that the database contains sequences of itemsets. But for the sake of simplicity, we will restrict patterns to sequences of items (each itemset is a singleton). Listing \ref{list:sequences_IS} in Appendices gives an encoding for the full general case of sequential pattern mining.

Our proposal is borrowed from J\"arvisalo's \cite{jarvisalo2011itemset}: the solution of the sequential pattern mining ASP program is all the answer sets (AS), each of which contains the atoms describing a single frequent pattern as well as its occurrences in database sequences. The solution relies on the ``generate and test principle'': generate combinatorially all the possible patterns and their related occurrences in the database sequences and test whether they satisfy the specified constraints.

\subsection{Modelling database, patterns and parameters}
\label{subseq:modelling}
A sequence database $\mathcal{D}$ is modelled by the predicate \lstinline!seq(T,Is,I)! which holds if sequence \lstinline!T! contains item \lstinline!I! at index \lstinline!Is!.

\begin{example}[A sequence database encoded in ASP]
The following facts encode the database of Example \ref{ex:seq_mining}:
\begin{lstlisting}[numbers=none]
seq(1,1,a). seq(1,2,c).
seq(2,1,d). seq(2,2,a). seq(2,3,b). seq(2,4,c).  
seq(3,1,b).
seq(4,1,a). seq(4,2,b). seq(4,3,c).
seq(5,1,a). seq(5,2,b).
seq(6,1,a). seq(6,2,c). seq(6,3,b). seq(6,4,c).
seq(7,1,a). seq(7,2,b). seq(7,3,c).
\end{lstlisting}
\end{example}

Similarly, the current pattern is modelled by the predicate \lstinline!pat(Ip,I)! which holds if the current pattern contains item \lstinline!I! at index \lstinline!Ip!. 

For example, the pattern $\langle a b c \rangle$ is modelled by the following atoms: \\ 
\noindent\lstinline!pat(1,a). pat(2,b). pat(3,c).!

In addition, we define two program constants:
\begin{itemize}
\item \lstinline!#const th=23.!
represents $f_min$, the minimal frequency threshold, \ie the requested minimal number of supporting sequences
\item \lstinline!#const maxlen=10.!
represents the maximal pattern length.
\end{itemize}

Let $S$ be a set of ground atoms and $P \subset S$ the set of \lstinline!pat(Ip,I)! atoms in $S$, according to the J\"arvisalo's encoding principle we would like an ASP program $\pi$ such that $S$ is an answer set of $\pi$ iff the pattern defined by $P$ is a frequent sequential pattern in the database $\mathcal{D}$.

\subsection{Two encodings for sequential pattern mining}
The main difficulty in declarative sequential pattern mining is to decide whether a pattern $P=\langle p_1, \dots, p_n\rangle$ supports a sequence $S=\langle s_1, \dots, s_m\rangle$ of the database. 
According to Def. \ref{def:insertableitem}, it means that it exists a mapping $e=(e_i)_{1\leq i \leq n}$ such that $p_i=s_{e_i}$. Unfortunately, this definition is not usable in practice to implement an efficient ASP encodings. The difficulty comes from the possible multiple mappings of a pattern in a single sequence. On the other hand, the detailed mapping description is not required here, we simply have to defined embeddings that exists iff a pattern supports a sequence.
An embedding of a pattern in a sequence is given by the description of a relation between pattern item indexes to sequence item indexes. 

\begin{figure}[tp]
\centering
\includegraphics[width=\textwidth]{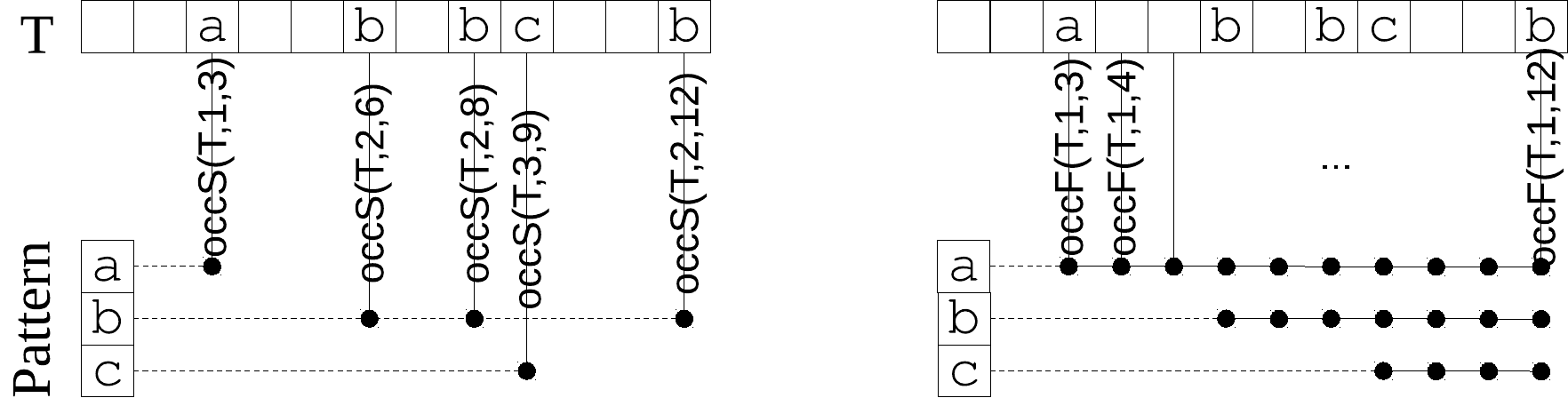}
\caption{Illustration of embeddings strategies. On the left, \emph{skip-gaps}, on the right, \emph{fill-gaps}.
}
\label{fig:embeddings}
\end{figure}

This section presents two encodings of sequential pattern mining. These two encodings differ in their representation of embeddings, as illustrated in Fig. \ref{fig:embeddings}. 
Two embedding strategies have been defined and compared in our results: \emph{skip-gaps} and \emph{fill-gaps}.

\espace

More formally, let $P=\langle p_1, \dots, p_n\rangle$ be a pattern sequence and $S_T=\langle s_1, \dots, s_m\rangle$ be the $T$-th sequence of $\mathcal{D}$.
In the \emph{skip-gaps} strategy, an embedding $\mathcal{E}$ is a relation over $[1,m]\times[1,n]$ such that $\forall (i,j) \in \mathcal{E},\; i \leq j \wedge p_i = s_j$ and $\forall (i, j), (i', j') \in \mathcal{E},\; i < i' \implies j < j'$. 
In the \emph{fill-gaps} strategy, an embedding $\mathcal{E'}$ is the same relation as $\mathcal{E}$ (i.e. $(i,j)\in\mathcal{E} \implies (i,j)\in\mathcal{E}'$) with the additional specification: $\forall i \in [1,m],\; \forall j,j' \in [1,n],\; (i, j)\in \mathcal{E'} \wedge j<j' \implies (i, j')\in \mathcal{E'}$. This additional specification expresses that once a pattern item has been mapped to the leftmost (having the lowest index, let it be $j$), the knowledge of this mapping is maintained on remaining sequences items with indexes $j'>j$. So, a \emph{fill-gaps} embedding makes only explicit the ``leftmost admissible matches'' of $P$ items in sequence $S_T$.

Relations $\mathcal{E}$ and $\mathcal{E}'$ are interesting because (i) that can be computed in a constructive way (\ie without encoding guesses) and (ii) they contains the information required to decide whether the pattern supports the sequence.

The two following sections detail the ASP programs for extracting patterns under each embedding strategy. 

\subsubsection{The skip-gaps approach}\label{sec:aspmining_skipgap}
In the first ASP encoding, an embedding of the current pattern $P=\langle p_i\rangle_{1\leq i \leq n}$ in sequence $T=\langle s_i\rangle_{1\leq i \leq m}$ is described by a set of atoms \lstinline!occS(T,Ip,Is)! 
which holds if \lstinline!Ip!-th the pattern item (occurring at index \lstinline!Ip!) is identical to the \lstinline!Is!-th item in sequence \lstinline!T! (formally, $p_\mathtt{Ip}=s_\mathtt{Is} \wedge \langle p_i\rangle_{1\leq i \leq \mathtt{Ip}}\prec \langle s_i\rangle_{1\leq i \leq \mathtt{Is}}$).
The set of valid atoms \lstinline!occS(T,_,_)! 
encodes the relation $\mathcal{E}$ above and is illustrated in Fig. \ref{fig:embeddings} (on the left).

\begin{example}[Illustration of skip-gaps embedding approach]\label{ex:skipgap}
Let $P=\langle a c \rangle$ be a pattern represented by \lstinline!pat(1,a).pat(2,c).! Here follows, the embeddings of pattern $P$ in the sequences of example \ref{ex:seq_mining}:
\begin{lstlisting}[numbers=none]
occS(1,1,1) occS(1,2,2)
occS(2,1,2) occS(2,2,4)
occS(4,1,1) occS(4,2,3)
occS(5,1,1)
occS(6,1,1) occS(6,2,2) occS(6,2,4)
occS(7,1,1) occS(7,2,3) 
\end{lstlisting}
The pattern could not be fully identified in the fifth sequence. There are two possible embeddings in the sixth sequence. Atom \lstinline!occS(6,1,1)! is used for both.
Nonetheless, this sequence must be counted only once in the support.
\end{example}

Listing \ref{list:sequences_sg} gives the ASP program for sequential pattern mining.  The first line of the program is called a \emph{projection}. It defines a new predicate that provides all items from the database. The symbol ``\lstinline!_!" denotes an anonymous (or don't care) variable.

\begin{lstlisting}[caption={Encoding of frequent sequential pattern mining -- skip-gaps strategy}, label=list:sequences_sg, float]
item(I) :- seq(_, _,I).

%sequential pattern generation
patpos(1).
0 { patpos(Ip+1) } 1 :- patpos(Ip), Ip<maxlen.
patlen(L) :- patpos(L), not patpos(L+1).

1 { pat(Ip,I): item(I) } 1 :- patpos(Ip).

%pattern embeddings
occS(T,1   ,Is):- seq(T,Is,I), pat(1,   I).
occS(T,Ip+1,Is):- seq(T,Is,I), pat(Ip+1,I), occS(T,Ip,Js), Js<Is.

%frequency constraint
support(T) :- occS(T,L,_), patlen(L).
:- { support(T) } < th.
\end{lstlisting}

Lines 4 to 8 of the program encode the pattern generation. Predicate \lstinline!patpos/1! defines the allowed sequential pattern indexes, beginning at index 1 (line 4). Line 5 is a choice rule that generates the successive pattern positions up to an ending index iterating from 2 to \lstinline!maxlen!: \lstinline!patpos(Ip+1)! is true if there is a pattern position at index \lstinline!Ip! and \lstinline!Ip! is lower than \lstinline!maxlen!.
Line 6 defines the length of a pattern: \lstinline!patlen(L)! holds if \lstinline!L! is the index of the last pattern item (there is no pattern item with a greater index). This predicate is used to decide whether an embedding has been completed or not. 
Finally, line 8 is a choice rule that associates exactly one item with each position \lstinline!X!. 
We can note that each possible sequence is generated once and only once. So, there is no redundancy in the search space exploration.

Lines 11 to 12 encode pattern embedding search. 
Line 11 guesses a sequence index for the first pattern item: \lstinline!occS(T,1,Is)! holds if the first pattern item is identical to the \lstinline!Is!-th of sequence \lstinline!T! (i.e. $p_1=s_\mathtt{Is}$).
Line 12 guesses sequence indexes for pattern items at indexes strictly greater than 1. \lstinline!occS(T,Ip,Is)! holds if the \lstinline!Ip!-th pattern item is equal to the \lstinline!Is!-th sequence item (i.e. $p_\mathtt{Ip}=s_\mathtt{Is}$) and the preceding pattern item is mapped to a sequence item at an index strictly lower than \lstinline!Is!. Formally, this rule expresses the following implication $(\mathtt{Jp},\mathtt{Is}-1) \in \mathcal{E} \wedge p_\mathtt{Ip}=s_\mathtt{Is}\wedge \mathtt{Ip}>\mathtt{Jp} \implies (\mathtt{Ip},\mathtt{Is}) \in \mathcal{E}$ and recursively, we have $\langle p_i\rangle_{1\leq i \leq \mathtt{Ip}}\prec \langle s_i\rangle_{1\leq i \leq \mathtt{Is}}$.
It should be noted that this encoding generates all the possible embeddings of some pattern.

Finally, lines 15 to 16 are dedicated to assess the pattern frequency constraint. \lstinline!support(T)! holds if the database sequence \lstinline!T! supports the pattern, \ie if an atom \lstinline!occS! holds for the last pattern position. The last line of the program is an integrity constraint ensuring that the number of supported sequences is not lower than the threshold \lstinline!th! or, in other words, that the support of the current pattern is greater than or equal to the threshold.

\subsubsection{The fill-gaps approach}\label{sec:aspmining_fillgap}
In the \emph{fill-gap} approach, an embedding of the current pattern $P$ is described by a set of atoms \lstinline!occF(T,Ip,Is)! having a slightly different semantics than in the \emph{skip-gap} approach.
\lstinline!occF(T,Ip,Is)! holds if at sequence index \lstinline!Is! it is true that the \lstinline!Ip!-th pattern item has been mapped (to some sequence index equal to \lstinline!Is! or lower than \lstinline!Is! if \lstinline!occF(T,Ip,Is-1)! holds). More formally, we have $\langle p_i\rangle_{1\leq i \leq \mathtt{Ip}}\prec \langle s_i\rangle_{1\leq i \leq \mathtt{Is}}$. 
The set of atoms \lstinline!occF(T,_,_)! encodes the relation $\mathcal{E'}$ above and is illustrated in Fig. \ref{fig:embeddings} (on the right). 

\begin{example}[Fill-gaps approach embedding example]\label{ex:fillgap}
Pattern $P=\langle a, c \rangle$ has the following fill-gaps embeddings (represented by \lstinline!occF! atoms) in the sequences of the database of example \ref{ex:seq_mining}:
\begin{lstlisting}[numbers=none]
occF(1,1,1) occF(1,1,2) occF(1,2,2) 
occF(2,1,2) occF(2,1,3) occF(2,1,4) occF(2,2,4)
occF(4,1,1) occF(4,1,2) occF(4,1,3) occF(4,2,3) 
occF(5,1,1) occF(5,1,2) 
occF(6,1,1) occF(6,1,2) occF(6,1,3) occF(6,1,4) occF(6,2,2) occF(6,2,3) occF(6,2,4)
occF(7,1,1) occF(7,1,2) occF(7,1,3) occF(7,2,3)
\end{lstlisting}

Contrary to the skip-gap approach example (see Example \ref{ex:skipgap}), the set of \lstinline!occF(T,Ip,Is)! atoms alone is not sufficient to deduce all occurrences. For instance, occurrence with indexes $(3,8,9)$ is masked.
\end{example}

\begin{lstlisting}[caption={Encoding of frequent sequential pattern mining -- fill-gaps strategy (see Listing \ref{list:sequences_sg} for first lines).}, label=list:sequences_fg, float,firstnumber=10]
%pattern embeddings
occF(T,1,Is)  :- seq(T,Is,I), pat(1,I).
occF(T,Ip,Is) :- occF(T, Ip-1, Is-1), seq(T,Is,I), pat(L,I).
occF(T,Ip,Is) :- occF(T, Ip,   Is-1), seq(T,Is,_).

%frequency constraint
seqlen(T,L) :- seq(T,L,_), not seq(T,L+1,_).
support(T) :- occF(T, L, LS), patlen(L), seqlen(T,LS).
:- { support(T) } < th.
\end{lstlisting}

Listing \ref{list:sequences_fg} gives the ASP program for sequential pattern mining with the fill-gaps strategy.
The rules are quite similar to those encoding the skip-gaps method. The main difference comes from the computation of embeddings (lines 11-13).
As in listing \ref{list:sequences_sg}, line 11 guesses a sequence index for the first pattern item: \lstinline!occF(T,1,Is)! holds if the first pattern item is identical to the \lstinline!Is!-th of sequence \lstinline!T! (i.e. $p_\mathtt{Ip}=s_\mathtt{Is}$).

Line 12 guesses sequence indexes for pattern items at indexes strictly greater than 1. \lstinline!occS(T,Ip,Is)! holds if the \lstinline!Ip!-th pattern item is equal to the \lstinline!Is!-th sequence item  and the preceding pattern item is mapped to some sequence item at some index strictly lower than \lstinline!Is!.  More formally, we have that $p_\mathtt{Ip}=s_\mathtt{Is} \wedge (\mathtt{Ip}-1,\mathtt{Is}-1) \in \mathcal{E}' \implies (\mathtt{Ip},\mathtt{Is}) \in \mathcal{E}'$. 

Line 13 simply maintains the knowledge that the \lstinline!Ip!-th pattern item has been mapped all along the further sequence indexes, i.e. \lstinline!occF(T,Ip,Is)! holds if \lstinline!occF(T,Ip,Is-1)! holds. More formally, $(\mathtt{Ip}-1,\mathtt{Is}-1) \in \mathcal{E}' \implies (\mathtt{Ip},\mathtt{Is}) \in \mathcal{E}'$. In combination with previous rules, we thus have recursively that \lstinline!occF(T,Ip,Is)! is equivalent to $\langle p_i\rangle_{1\leq i \leq \mathtt{Ip}}\prec \langle s_i\rangle_{1\leq i \leq \mathtt{Is}}$.

Line 17 a sequence is supported by the pattern an \lstinline!occF! atoms exists at the last position \lstinline!LS! of the sequence, computed line 16.
The remaining rules for testing whether it is greater than the threshold  \lstinline!th! are identical to those in the skip-gaps approach.

\subsection{Sequential pattern mining improvements}
The main objective of this subsection is to present alternative encodings of the sequential pattern mining task. These encodings attempt to take advantage of known properties of the sequential pattern mining task to support the solver to mine datasets more efficiently or with less memory requirements. The efficiency of these improvements will be compared in the experimental part.

\subsubsection{Filter out unfrequent items}
The first improvement consists in generating patterns from only frequent items. According to the anti-monotonicity property, all items in a pattern have to be frequent.
The rules in listing \ref{list:frequentitems} may replace the projection rule previously defining the available items. Instead, an explicit aggregate argument is introduced to evaluate the frequency of each item \lstinline!I! and to prune it if it is unfrequent.

In the new encoding, the predicate \lstinline!sitem/1! defines the set of items that occurs in the database and \lstinline!item/1! defines the frequent items that can generate patterns.

\begin{lstlisting}[caption={Restriction of the pattern traversal to sequences made of frequent items only}, label=list:frequentitems]
sitem(I) :- seq(_, _,I).
item(I)  :- sitem(I), #count{ T:seq(T,_,I) } >= th.
\end{lstlisting}

\subsubsection{Using projected databases}
The idea of this alternative encoding is to use the principle of \emph{projected databases} introduced by algorithm PrefixSpan \cite{Pei2004}. Let $P=\langle p_1, \dots, p_n \rangle$ be a pattern, the projected database of $\mathcal{D}=\{S_1, \dots, S_n\}$ is $\{S'_1, \dots, S'_n\}$ where $S'_i$ is the projected sequence of $S_i$ with respect to $P$. Let $S_i=\langle s_1,\dots, s_m\rangle$ be a sequence. Then the projected sequence of $S_i$ is $S'_i=\langle s_{k+1},\dots s_m\rangle$ where $k$ is the position of the last item of the first occurrence of $P$ in $S_i$. If $P$ does not occur in $S_i$ then $S'_i$ is empty.

A projected database is smaller than the whole database and the set of its frequent items is consequently much smaller than the original set of frequent items. The idea is to improve the candidate generation part of the algorithm by making use of items from projected databases. Instead of generating a  candidate (a sequential pattern) by extending a frequent pattern with an item that is frequent in the whole database, the pattern extension operation uses only the items that are frequent in the database projected along this pattern.

\begin{lstlisting}[caption={Pattern generation with prefix-projection principle -- skip-gaps strategy}, label=list:seq_proj_sg, firstnumber=8]
item(1,I)   :- sitem(I),
               #count{ T: seq(T,_,I) } >= th.
item(Ip+1,I):- item(Ip,I),
               #count{ T: seq(T,Js,I),occS(T,Ip,Is),Js>Is }>= th.
1 { pat(Ip,I) : item(Ip,I) } 1 :- patpos(Ip).
\end{lstlisting}

\begin{lstlisting}[caption={Pattern generation with prefix-projection principle -- fill-gaps strategy}, label=list:seq_proj_fg, firstnumber=8]
item(1,I)    :- sitem(I),
                #count{ T: seq(T,_,I) } >= th.
item(Ip+1,I) :- item(Ip,I),
                #count{ T: seq(T,Is,I), occF(T,Ip,Is) } >= th.
1 { pat(Ip,I) : item(Ip,I) } 1 :- patpos(Ip).
\end{lstlisting}

The ASP encoding of the prefix-projection principle is given in Listing \ref{list:seq_proj_sg} for the skip-gaps strategy and in Listing \ref{list:seq_proj_fg} for the fill-gaps strategy. 
The programs of Listings \ref{list:sequences_sg} and \ref{list:sequences_fg} remain the same except for the generation of patterns defined by \lstinline!patpos/1! and the new predicate \lstinline!item/2!. \lstinline!item(Ip,I)! defines an item \lstinline!I! that is frequent in sequence suffixes remaining after removing the prefix of the sequence containing the first occurrence of the \lstinline!X-1!-pattern prefix (consisting of the \lstinline!Ip-1! first positions of the pattern). 
Lines 8-9 are similar to those in Listing \ref{list:frequentitems}. 
\lstinline!item(1,I)! defines the frequent items, \ie those that are admissible as first item of a frequent pattern. Lines 10-11 generates the admissible items for pattern position \lstinline!Ip+1!. Such an item must be admissible for position \lstinline!Ip! and be frequent in sequence suffixes (sub-sequence after at least one (prefix) pattern embedding). For  skip-gaps, the sequence suffix is defined by \lstinline!seq(T,Js,I), occS(T,Ip,Is), Js>Is! (the items at sequence positions farther away than the last position that matches the last (partial) pattern item at position \lstinline!Ip!). For fill-gaps, \lstinline!seq(T,Js,I), occF(T,Ip,Is)! is sufficient because \lstinline!occF(T,Ip,Is)! atoms represent the sequence suffix beginning at the sequence position that matches the last (partial) pattern item (at position \lstinline!Ip!).

\subsubsection{Mixing itemsets and sequences mining}
In \cite{jarvisalo2011itemset}, J\"{a}rvisalo showed that ASP can be efficient for itemset pattern mining. The main idea of this last alternative approach is to mine frequent itemsets and to derive sequential patterns from them.

This time, the itemset mining step extracts a frequent itemset pattern $I=(e_i)_{i\in[n]}$, $e_i \in \mathcal{I}$. A sequential pattern $S=(s_i)_{i\in[m]}$ is generated using the items of the itemset, \ie $\forall i\in[m]$, $\exists j\in[n],\, s_i=e_j$ taking into account that items may be repeated within a sequential pattern and that every item from $I$ must appear in $S$. If not, there would exist a subset $J\subset I$ that would generate the same sequence $s$. This would lead to numerous redundant answer sets for similar frequent sequences and would cause a performance drop.

\begin{lstlisting}[caption={Mining frequent sequences from frequent itemsets -- skip-gaps strategy}, label=list:sequences_itemsets,float]
sitem(I) :- seq(_, _,I).

% extract frequent itemsets
db(T,I) :- seq(T,_,I).
0 { in_itemset(I) } 1 :- th { in_support(T):db(T,I) }, sitem(I).
in_support(T) :- 0 { conflict_at(T,I) : sitem(I) } 0, db(T,_).
conflict_at(T,I) :- not db(T,I), in_itemset(I), db(T,_).
               
%sequential pattern generation from frequent itemsets
patpos(1).
0 { patpos(Ip+1) } 1 :- patpos(Ip), Ip<maxlen.
patlen(L) :- patpos(L), not patpos(L+1).

1 { pat(Ip,I) : in_itemset(I) } 1 :- patpos(Ip).

:- 0 { pat(Ip,I) : patpos(Ip) } 0 , in_itemset(I).

%pattern occurrences
occS(T,1,Is)   :- in_support(T), seq(T,Is,I), pat(1,   I).
occS(T,Ip+1,Is):- occS(T,Ip,Js), seq(T,Is,I), pat(Ip+1,I), Js<Is.

support(T) :- occS(T, L, _), patlen(L).
:- { support(T) } < th.

\end{lstlisting}

Listing \ref{list:sequences_itemsets} gives the entire encoding of this alternative for the skip-gaps strategy\footnote{A similar encoding can be done for the fill-gaps strategy applying the same changes as above.}. 
Rules in Lines 4-7 extract frequent itemsets, represented by the predicate \lstinline!in_itemset/1!, borrowed from J\"{a}rvisalo's encoding \cite{jarvisalo2011itemset}. 
Next, the generation of sequential patterns in line 14 uses only items from such a frequent itemset. Line 16 defines a constraint required to avoid answer set redundancies. The remaining part of the program is left unchanged.

\section{Alternative sequential pattern mining tasks}\label{sec:alternative_spm_asp}
In this section, we illustrate how the previous encodings can be modified to solve more complex mining tasks. Our objective is to show the impressive expressiveness of ASP which let us encode a very wide range of mining tasks. 
We focus our attention on the most classical alternative sequential pattern mining tasks: constrained sequential patterns and condensed representation of sequential patterns. 

In \cite{Negrevergne15}, the authors organize the constraints on sequential patterns in three categories: 1) constraints on patterns, 2) constraints on patterns embeddings, 3) constraints on pattern sets. These constraints are provided by the user and capture his background knowledge.

The following subsection shows that our ASP approach enables to add constraints on individual patterns (constraints of categories 1 and 2). But, as ASP cannot compare models with each others, the third category of constraints can not be encoded directly. 

In Sect. \ref{sec:condensed}, we transform the classical definition of the most known constraints of the third category -- the condensed representations -- to encode them in pure ASP. 
Condensed representations (maximal and closed patterns) have been widely studied due to their monotonicity property, and to their representativeness with respect to frequent patterns.
Concerning more generic constraints on pattern sets, such as the extraction of skypatterns \cite{ugarte2015skypattern}, we have proposed in \cite{Guyet_IJCAI2016} an ASP-based approach for mining sequential skypatterns using \asprin\ for expressing preferences on answer sets. \asprin\ \cite{brewka:hal-01187001} provides a generic framework for implementing a broad range of preferences relations on ASP models and can easily manage them. This approach is out of the scope of this article.

\subsection{Constraints on patterns and embeddings}
Pei et \al \cite{pei2007seqmining} defined seven types of constraints on patterns and embeddings. In this subsection, we describe each of these constraints keeping their original numbering. Constraints 1, 2, 3 and 5 are pattern constraints, while constraints 4, 6 and 7 are embedding constraints.
If not stated otherwise, the base encoding is the skip-gaps strategy and line numbers refers to Listing \ref{list:sequences_sg}. 

In a first approach, constraints on patterns and on embeddings may be trivially encoded by adding integrity constraints. But these integrity constraints acts a posteriori, during the test stage, for invalidating candidate models. A more efficient method consists in introducing constraints in the generate stage, specifically in choice rules, for pruning the search space early. 

\textbf{Constraint 1 -- item constraint.}
An item constraint specifies what are the particular individual or groups of items that should or should not be present in the patterns. For instance, the constraint ``patterns must contain item \lstinline!1! but not item \lstinline!2! nor item \lstinline!3!'' can be encoded using \lstinline!must_have/1! and \lstinline!cannot_have/1! predicates: \lstinline!must_have(1). cannot_have(2). cannot_have(3).! 

To cope with this kind of constraint, Line 8 of Listing \ref{list:sequences_sg} is modified as:
\begin{lstlisting}[firstnumber=8]
1 { pat(X,I): item(I), not cannot_have(I) } 1 :- patpos(X).
:- { pat(X,I) : must_have(I) } < 1.
\end{lstlisting}

The encoding of Line 8 modifies the choice rule to avoid the generation of known invalid patterns, \ie patterns with forbidden items.
Line 9 is a new constraint that imposes to have at least one of the required items.

\textbf{Constraint 2 -- length constraint.}
A length constraint specifies a prerequisite on pattern length. The maximal length constraint is anti-monotonic while the minimal length is not anti-monotonic. The maximal length constraint is already encoded using the program constant \lstinline!maxlen! in our encodings. A new constant \lstinline!minlen! is defined to encode the minimal length constraint and a new rule is added to predicate \lstinline!patpos/1! to impose at least \lstinline!minlen! positions in patterns instead of only one.
\begin{lstlisting}[numbers=none]
#const minlen = 2.
patpos(1).    
patpos(X+1) :- patpos(X), X<=minlen.
0 { patpos(X+1) } 1 :- patpos(X), X<maxlen.
\end{lstlisting}

\textbf{Constraint 3 -- super-pattern constraint.}
A super-pattern constraint enforces the extraction of patterns that contain one or more given sub-patterns.
Mandatory sub-patterns are defined by means of the new predicate \lstinline!subpat(SP,P,I)! expressing that sub-pattern \lstinline!SP! contains item \lstinline!I! at position \lstinline!P!.

Predicate \lstinline!issubpat(SP)! verifies that the sub-pattern \lstinline!SP! is included in the pattern. An approach similar to embedding computation may be used:
\begin{lstlisting}[numbers=none]
issubpat(SP,1,P) :- pat(P,I), subpat(SP,1,I).
issubpat(SP,Pos+1,P) :- issubpat(SP,Pos,Q), pat(P,I),
                        subpat(SP,Pos+1,I), Q<P.
issubpat(SP) :- issubpat(SP,L,_), subpatlen(SP,L).
\end{lstlisting}

\lstinline!issubpat(SP)! is true if the sub-pattern \lstinline!SP! is a sub-pattern of the current pattern. This predicate is used to define the final integrity constraint:
\begin{lstlisting}[numbers=none]
:- #count{ SP : issubpat(SP), subpat(SP,_,_) } = 0.
\end{lstlisting}

\textbf{Constraint 4 -- aggregate constraint.}
An aggregate constraint is a constraint on an aggregation of items in a pattern, where the aggregate function can be \emph{sum}, \emph{avg}, \emph{max}, \emph{min}, \emph{standard deviation}, etc.
The only aggregates that are provided by \clingo\ are \lstinline!#sum!, \lstinline!#max! and \lstinline!#min!.
For example, let us assume that to each item \lstinline!I! is assigned a cost \lstinline!C!, which is given by predicate \lstinline!cost(I,C)!. The following constraint enforces the selection of patterns having a total cost of at least 1000.
\begin{lstlisting}[numbers=none]
:- #sum{ C,X : cost(I,C), pat(X,I) } < 1000.
\end{lstlisting}
As an integrity constraint, this rule means that it is not possible to have a total amount lower than 1000 for pattern. It should be noted that \lstinline!C! values are summed for each pair $(\texttt{C},\texttt{X})$. Thus, item repetitions are taken into account.

\textbf{Constraint 5 -- Regular expression.}
Such a constraint is satisfied if the pattern is an accepted regular expression as stated by the user. A regular expression can be encoded in ASP as its equivalent deterministic finite automata. Expressing such a constraint is mainly technical and is not detailed here. SPIRIT \cite{SPIRIT1999} is one of the rare algorithms that considers complex pattern constraints expressed as regular expressions.

\textbf{Constraint 6 -- Duration constraints.}
The duration (or span) of some pattern is the difference between its last item timestamp and its first item timestamp. A duration constraint requires that the pattern duration should be longer or shorter than a given time period. In the database encoding introduced Sect. \ref{subseq:modelling}, predicate \lstinline!seq(T,P,I)! defines the timestamp of \lstinline!I! in sequence \lstinline!T! as the integer position \lstinline!P!~. 
A global constraint such as \emph{max-span} cannot be expressed through simple local constraints on successive pattern item occurrences, as gap constraints described in the next paragraph. In fact, the predicate \lstinline!occS/3! does not describe the embeddings precisely enough to express the \emph{max-span} constraint: for some pattern embedding, there is no explicit link between its first item occurrence and its last item occurrence.
The proposed solution is to add an argument to \lstinline!occS/3! to denote the position of the occurrence of the first pattern item:
\begin{lstlisting}[firstnumber=11]
%pattern embeddings (skip-gaps strategy)
occS(T,1,P,P)      :- seq(T,P,I), pat(1,I).
occS(T,Pos+1,P,IP) :- occS(T,Pos,Q,IP), seq(T,P,I), pat(Pos+1,I),
                      P-IP+1<=maxspan, P-IP+1>=minspan.
\end{lstlisting}

\textbf{Constraint 7 -- Gap constraints.}
A gap constraint specifies the maximal/minimal number of positions (or timestamp difference) between two successive itemsets in an embedding. The maximal gap constraint is anti-monotonic while the minimal gap is not anti-monotonic.
Contrary to pattern constraints, embedding constraints cannot be encoded simply by integrity constraints. In fact, an integrity constraint imposes a constraint on all embeddings. If an embedding does not satisfy the constraint, the whole interpretation -- \ie the pattern -- is unsatisfied.

In the following we give an encoding for the \emph{max-gap} and \emph{min-gap} constraints. For such local constraint, the solution consists in modifying the embedding generation (lines 11-12 in Listing \ref{list:sequences_sg}) for yielding only embeddings that satisfy gap constraints:
\begin{lstlisting}[firstnumber=11]
occS(T,1,P)     :- seq(T,P,I), pat(1,I).
occS(T,Pos+1,P) :- seq(T,P,I), pat(Pos+1,I), occS(T,Pos,Q),
                   P-Q-1>=mingap, P-Q-1<=maxgap.
\end{lstlisting}
This encoding assumes that the value of constants \lstinline!mingap! and \lstinline!maxgap! have been provided by the user (using \lstinline!#const! statements).

\espace

Constraints of type 6 and 7 can be mixed by merging the two encodings of \lstinline!occS! above:
\begin{lstlisting}[firstnumber=11]
occS(T,1,P,P)      :- seq(T,P,I), pat(1,I).
occS(T,Pos+1,P,IP) :- seq(T,P,I), pat(Pos+1,I), occS(T,Pos,Q,IP),
                                P-Q-1>=mingap, P-Q-1<=maxgap,
                                P-IP+1<=maxspan, P-IP+1>=minspan.
\end{lstlisting}

\subsection{Condensed representation of patterns: closed and maximal sequences} \label{sec:condensed}
In this section, we study the encodings for two well-studied pattern types, closed and maximal patterns. A closed pattern is such that none of its frequent super-patterns has the same support. A maximal pattern is such that none of its super-patterns is frequent. Thus, it is necessary to compare the supports of several distinct patterns. Since a solution pattern is encoded through an answer set, a simple solution would be to put constraints on sets of answer sets.
However, such a facility is not provided by basic ASP language\footnote{\asprin\ \cite{brewka:hal-01187001} is a \clingo\ extension that allows for this kind of comparison. For more details about the use of \asprin \ to extract skypatterns, see \cite{Guyet_IJCAI2016}.}. 
So, these constraints have been encoded without any comparison of answer sets but as additional constraints on the requested patterns.
The next section introduces the definitions of these alternative mining tasks and the properties that were used to transform the pattern set constraints as constraints on individual patterns. Sect. \ref{sec:maximal_patterns} gives encodings for closed and maximal patterns extraction.

\subsubsection{Definitions and properties}
A frequent pattern $S$ is \emph{maximal} (resp. \emph{backward-maximal}) with respect to the relation $\prec$ (resp. $\prec_b$) iff there is no other frequent pattern $S'$ such that $S \prec S'$ (resp. $S \prec_b S'$).

A frequent pattern $S$ is \emph{closed} (resp. \emph{backward-closed}) with respect to the relation $\prec$ (resp. $\prec_b$) iff there is no proper superpattern $S'$ such that $S \prec S'$ (resp. $S \prec_b S'$) and $supp(S)=supp(S')$.
Mining the closed patterns significantly reduces the number of patterns without loss of information for the analyst. 
Having the closed patterns and their support, the support of any pattern can be computed. This is not the case for maximal patterns.

\begin{example}[Maximal and closed-sequential pattern mining]\label{ex:closed_mining}
Considering the data\-base of Example \ref{ex:seq_mining}, among the frequent patterns with $f_{min}=3$, the only maximal pattern is $\langle abc\rangle$. The set of backward-maximal is $\{ \langle c\rangle, \langle bc\rangle, \langle ac\rangle, \langle abc\rangle\}$.

The set of closed patterns is $\{ \langle a \rangle, \langle b \rangle, \langle ab \rangle, \langle ac \rangle, \langle abc \rangle\}$. $\langle bc \rangle$ is not closed because in any sequence it occurs, it is preceded by an $a$. Thus $supp(\langle bc \rangle)=supp(\langle abc \rangle)=4$.

The set of backward-closed patterns is $\{ \langle a  \rangle, \langle b \rangle, \langle c \rangle, \langle bc \rangle, \langle ac \rangle, \langle abc \rangle\}$. $\langle bc\rangle$ is backward-closed because any pattern $\langle bc?\rangle$ is frequent.
\end{example}

Now, we introduce alternative maximality/closure conditions.
The objective of these equivalent conditions is to define maximality/closure without comparing patterns. Such conditions can be used to encode the mining of condensed pattern representations.
The main idea is to say that a sequence $S$ is maximal (resp. closed) if and only if for every sequence $S'$ s.t. $S$ is a subsequence of $S'$ with $|S'|=|S|+1$, then $S'$ is not frequent (resp. $S'$ has not the same support as $S$).

More precisely, a frequent pattern $S$ is maximal iff any sequence $S^{j}_{a}$, obtained by adding to $S$ any item $a$ at any position $j$, is non frequent. Such an $a$ will be called an \emph{insertable} item.

\begin{prop}[Maximality condition]\label{prop:maximality}
A frequent sequence $S=\langle t_1,\dots, t_n\rangle$ is maximal iff $\forall a \in \mathcal{I}$, $\forall j
\in [0,n]$, $|\{T \in \mathcal{D}| S \prec T
\wedge S^{j}_{a}\prec T\}|<f_{min}$, where $S^{0}_{a}=\langle a, t_{1},\dots, t_n\rangle$, $S^{j}_{a}=
\langle t_1,\dots, t_{j}, a, t_{j+1},\dots, t_n\rangle$ and
$S^{n}_{a}=\langle t_{1},\dots, t_n, a\rangle$.
\end{prop}

A frequent pattern $S$ is closed iff for any frequent sequence $S^{j}_{a}$, obtained by adding any item $a$ at any position $j$ in $S$, any sequence $T$ that supports $S$ supports also $S^{j}_{a}$.

\begin{prop}[Closure condition]\label{prop:closure}
A frequent sequence $S=\langle t_1,\dots, t_n\rangle$ is closed iff 
$\forall a \in \mathcal{I}$, $\forall j \in
[0,n]$, $supp(S^{j}_{a}) \geq f_{min} \implies (\forall T \in \mathcal{D}, S \prec T \implies  S^{j}_{a}\prec T)$, where $S^{0}_{a}=
\langle a, t_{1},\dots, t_n\rangle$, $S^{j}_{a}=\langle t_1,\dots, t_{j}, a,
t_{j+1},\dots, t_n\rangle$ and $S^{n}_{a}=\langle t_{1},\dots,
t_n, a\rangle$.
\end{prop}

\espace

A consequence (the contraposition) of these properties is that if an item may be inserted between items of an embedding for at least $f_{min}$ sequences (resp. for all supported sequences) then the current pattern is not maximal (resp. not closed). The main idea of our encodings is grounded on this observation. 

\begin{figure}[tb]
\sidecaption
\centering
\includegraphics[width=.5\textwidth]{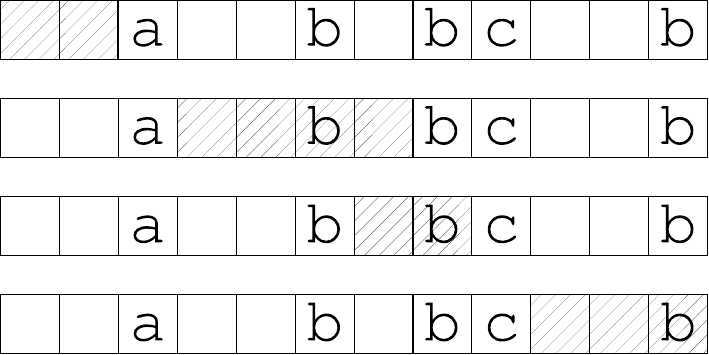}
\caption{Illustration of the notion of insertable region on the example of Fig. \ref{fig:embeddings} for pattern $\langle abc \rangle$. Each line shows an insertable region, from top to bottom: insertion in the prefix, insertion between $b$ and $b$, insertion in $b$ and $c$, insertion in the suffix.}
\label{fig:insertion_regions}
\end{figure}

The main difficulty is to construct the set of insertable items for each in-between position of a pattern, so-called insertable regions.
Fig. \ref{fig:insertion_regions} illustrates the insertable regions of a sequence for the pattern $\langle abc \rangle$. 

\begin{definition}[Insertable item/insertable region]\label{def:insertableitem}
Let $P=\langle p_i \rangle_{i\in[l]}$ be an $l$-pattern, $S=\langle s_i \rangle_{i\in[n]}$ be a sequence and $\epsilon^j=(e^j_i)_{i\in[l]}$, $j\in[k]$ be the $k$ embeddings of $P$ in $S$, $k>0$.
An \emph{insertable region} $R_i=[l_i+1,u_i-1]\subset [n]$, $i\in[l+1]$ is a set of positions in $S$ where $l_i \eqdef \min\limits_{j\in[k]}e^j_{i-1}, i\in [2,l+1], u_i \eqdef \max\limits_{j\in[k]}e^j_{i}, i\in[1,l])$, $l_1\eqdef 0$, $u_{l+1}\eqdef n+1$. 

Any item $a \in s_{p},$ $p\in [l_i,u_i]$, $i\in[l+1]$ is called an \emph{insertable item} and is such that $S$ supports the pattern $P'$ obtained by inserting $a$ in $P$ at position $i$ as follows:
\begin{itemize}
\item $P'=\langle a, p_{1},\dots, p_l\rangle$ if $i=1$,
\item $P'=\langle p_{1},\dots, p_l, a\rangle$ if $i=l+1$,
\item $P'=\langle p_{1},\dots, p_{i-1}, a, p_{i},\dots, p_l\rangle$ otherwise.
\end{itemize} 
\end{definition}

In the sequel, we present encodings for closed and maximal patterns which are based on the notations introduced in Definition \ref{def:insertableitem}. These encodings cope with the most general case of condensed patterns. It should be noted that, for  efficiency reasons, most of procedural algorithms for condensed sequential pattern mining process backward-condensed patterns only.
Specific ASP encodings for backward-condensed pattern mining can be found in \cite{Guyet_EGC2016}. These encodings are known to be more efficient but are less generic. In Sect. \ref{sec:experiments}, the performance of the encodings introduced here will be compared with other existing approaches that often implement only backward closure/maximality constraints.

\subsubsection{Encoding maximal and closed patterns constraints}\label{sec:maximal_patterns}

\begin{figure}[tb]
\includegraphics[width=5cm]{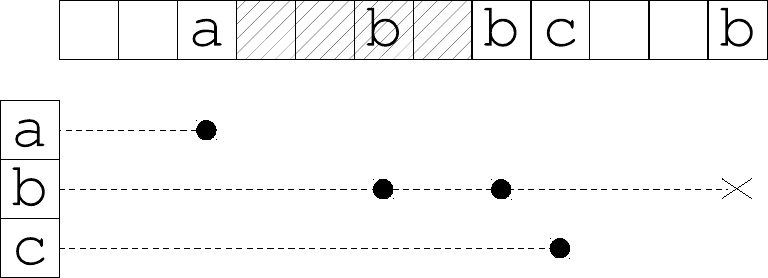} \hfill
\includegraphics[width=5cm]{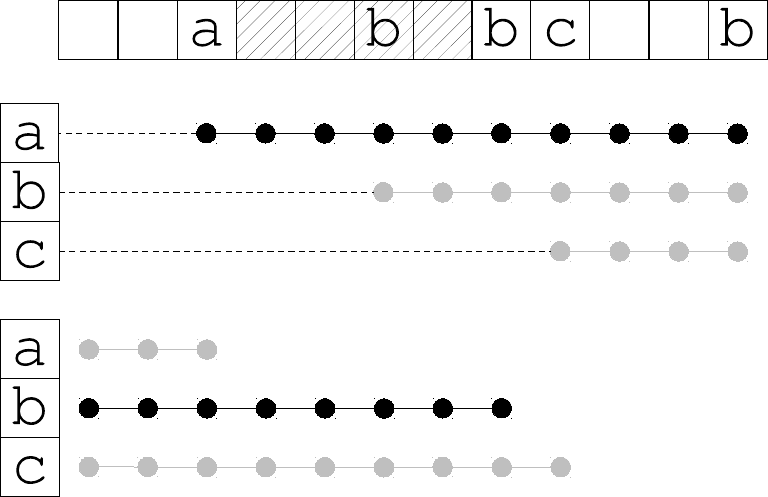}
\caption{Illustration of the computation of an insertable region (hatched area) where insertable items are located between $a$ and $b$ and related to the first and second element of pattern $\langle abc\rangle$.
On the left, valid \lstinline!occS/3! atoms in the skip-gaps strategy. In the figures, the leftmost occurrences and the rightmost occurrences are the same. Concerning the fill-gaps strategy, \lstinline!occF/3! and \lstinline!roccF/3! atoms are illustrated on the right. Black occurrences are used to compute the hatched region.}
\label{fig:insertion_sg}
\end{figure}

The encoding below describes how is defined the set of items that can be inserted between successive items of an embedding. These itemsets are encoded by the atoms of predicate \lstinline!ins(T,X,I)!  where \lstinline!I! is an item which can be inserted in an embedding of the current pattern in sequence \lstinline!T! between items at position \lstinline!X! and \lstinline!X+1! in the pattern.
We give the encodings for the two strategies skip-gaps and fill-gaps: Listing \ref{list:insertable_sg} (resp. Listing \ref{list:insertable_fg}) has to be added to the encoding of skip-gaps strategy (Listing \ref{list:sequences_sg}), resp. fill-gaps strategy (Listing \ref{list:sequences_fg}). We illustrate the way they proceed in Fig. \ref{fig:insertion_sg}.

\begin{lstlisting}[firstnumber=18, label={list:insertable_sg}, caption={Computation of insertable items -- skip-gaps strategy},float]
% leftmost "valid" embeddings
mlocc(T,1,P) :- occS(T,1,P), 0 { occS(T,1,Q): Q<P } 0, 
                support(T).
mlocc(T,X,P) :- occS(T,X,P), mlocc(T,X-1,Q), Q<P, X>1, 
                support(T).

% rightmost "valid" embeddings
mrocc(T,L,P) :- occS(T,L,P), 0 { occS(T,L,R): R>P } 0, patlen(L).
mrocc(T,X,P) :- occS(T,X,P), mrocc(T,X+1,R), R>P, X<L, patlen(L).

%insertable items
ins(T,1  ,I) :- seq(T,P,I), P<Q, mrocc(T,1,  Q).
ins(T,X  ,I) :- seq(T,P,I), P<Q, mrocc(T,X,  Q),
                            P>R, mlocc(T,X-1,R), X>1, patpos(X).
ins(T,L+1,I) :- seq(T,P,I), P>R, mlocc(T,L,  R), patlen(L).
\end{lstlisting}

\espace

Listing \ref{list:insertable_sg} gives the encoding for computing insertable items using the skip-gaps strategy. This encoding is based on the idea that the insertable region $i$ is roughly defined by the first occurrence of the $(i-1)$-th pattern item and the last occurrence of the $i$-th pattern item. 
However, not all occurrences of an item $I$ represented by \lstinline!occS/3! atoms are valid. For instance, in Fig. \ref{fig:insertion_sg}, on the left, the last occurrence of $b$ is not valid because it can not be used to define an occurrence of $\langle abc \rangle$. The valid occurrences are those which have both a preceding and a following valid occurrence. 
 Thus, this validity property is recursive.
The encoding of Listing \ref{list:insertable_sg} selects two types of occurrences: the leftmost occurrences (resp. rightmost occurrences) corresponding to the earlier (resp. the later) embeddings. 

Lines 19 and 25 are boundary cases. A leftmost occurrence is valid if it is the first occurrence in the sequence.
Lines 21-22 expresses that an occurrence of the $X$-th item is a valid leftmost occurrence if it follows a valid leftmost occurrence of the $(X-1)$-th item. Note that it is not required to compute a unique leftmost occurrence here.
Lines 25-26 do the same operation starting from the end of the sequence, precisely, the rightmost occurrence.

Lines 29-32 define insertable items. There are three cases. 
Lines 29 and 32 are specific boundary cases, \ie insertion respectively in the prefix and in the suffix. The rule in lines 30-31 specifies that insertable items \lstinline!I! are the items of a sequence \lstinline!T! at position \lstinline!P! such that \lstinline!P! is strictly between a leftmost position of the \lstinline!(X-1)!-th item and a rightmost position of the \lstinline!X!-th item. In Fig. \ref{fig:insertion_sg} left, the hatched segment defines the second insertable region for pattern $\langle abc \rangle$ (strictly between $a$ and $b$).

\espace

The encoding of Listing \ref{list:insertable_fg} achieves the same task using the alternative semantics for predicate \lstinline!occF/3! defining the fill-gaps strategy. As noted for the previous encoding, only the positions of the last and the first valid occurrences are required for any pattern item. It can be noted that the fill-gaps strategy provides the first valid occurrence of an item \lstinline!X! as the first atom of the \lstinline!occF(T,X,_)! sequence.
Then, computing the last occurrence for each pattern item can be done in the same manner considering an embedding represented in reverse order .
The right part of Fig. \ref{fig:insertion_sg} illustrates \lstinline!occF/3! and \lstinline!roccF/3! (reverse order) occurrences (see Listing \ref{list:insertable_fg}, lines 21-23).
We can notice that the hatched insertable region is the intersection of occurrences related to $a$ and reverse occurrences related to $b$, after having removed intersection bounds.

\begin{lstlisting}[firstnumber=20, label={list:insertable_fg}, caption={Computation of insertable items -- fill-gaps strategy},float]
%embeddings in a reverse order
roccF(T,L,P) :- seq(T,P,I), pat(L,I), patlen(L).
roccF(T,L,P) :- roccF(T, L,   P+1), seq(T,P,_).
roccF(T,L,P) :- roccF(T, L+1, P+1), seq(T,P,C), pat(L,C).

%insertable items
ins(T,1  ,I) :- seq(T,P,I), roccF(T,1,  P+1).
ins(T,L+1,I) :- seq(T,P,I), occF(T,L,P-1), patlen(L).
ins(T,X,  I) :- seq(T,P,I), roccF(T,X,P+1),
                            occF(T,X-1,P-1), patpos(X), X>1.
\end{lstlisting}

The computation of insertable items, Listing \ref{list:insertable_fg} lines 26-29, exploits the above remark. Line 26 defines the insertable region in a prefix using \lstinline!roccF(T,1,P)!. Since items are insertable if they are strictly before the first position, we consider the value of \lstinline!roccF(T,1,P+1)!. Line 27 uses \lstinline!occF(T,L,P)! to identifies the suffix region. Line 28-29 combines both constraints for in-between cases.

\espace

We can now define the (integrity) constraints for closed and maximal patterns. These constraints are the same for the two embedding strategies.

To extract only maximal patterns, the following constraint denies patterns for which it is possible to insert an item which will be frequent within sequences that support the current pattern.
\begin{lstlisting}[numbers=none]
:- item(I), X = 1..maxlen+1, { ins(T,X,I) : support(T) } >= th.
\end{lstlisting}

The following constraint concerns the extraction of closed-patterns. It specifies that for each insertion position (from 1, in the prefix, to \lstinline!maxlen!$+1$, in the suffix), it not possible to have a frequent insertable item \lstinline!I! for each supported transaction.
\begin{lstlisting}[numbers=none]
:- item(I), X = 1..maxlen+1, { ins(T,X,I) } >=th,
                             ins(T,X,I) : support(T).
\end{lstlisting}

\espace

Though interesting from a theoretical point of view, these encodings leads to more complex programs and should be more difficult to ground and to solve, especially the encoding in Listing \ref{list:insertable_sg}. Backward-closure/maximality constraints are more realistic from a practical point of view.

Finally, it is important to notice that condensed constraints have to be carefully combined with other patterns/embedding constraints. As noted by Negrevergne et \al \cite{negrevergne2013dominance}, in such cases the problem is not clearly specified. For instance, with our database of Example \ref{ex:seq_mining}, extracting closed patterns amongst the patterns of length at most 2 will not yield the same results as extracting closed patterns of length at most 2. In the first case, $\langle bc \rangle$ is closed because there is no extended pattern (of length at most 2) with the same support. In the second case, this pattern is not closed (see Example \ref{ex:closed_mining}), even if its length is at most 2.

\section{Related works}\label{sec:soa}
Sequential pattern mining in a sequence database have been addressed by numerous algorithms inspired by algorithms for mining frequent itemsets. The most known algorithms are GSP \cite{Srikant96}, SPIRIT \cite{SPIRIT1999}, SPADE \cite{Zaki01}, PrefixSpan \cite{Pei2004}, and CloSpan \cite{CloSpan03} or BIDE \cite{BIDE2004} for closed sequential patterns.
It is worth-noting that all these algorithms are based on the anti-monotonicity property which is essential to obtain good algorithmic performances. The anti-monotonicity property states that if some pattern is frequent then all its sub-patterns are also frequent. And reciprocally, if some pattern is not-frequent then all its super-patterns are non-frequent. This property enables the algorithm to prune efficiently the search space and thus reduces its exploration.
These algorithms differ by their strategy for browsing the search space. GSP \cite{Srikant96} is based on a breadth-first strategy, while PrefixSpan \cite{Pei2004} combines a depth-first strategy with a database projection that consists in reducing the database size after each pattern extension. LCM\_seq \cite{LCM_seq} is also based on the PrefixSpan principle, but it uses the data structures and processing method of LCM, which is the state of the art algorithm for frequent itemsets mining. Finally, SPADE \cite{Zaki01} introduces a vertical representation of database to propose an alternative to the two previous type of algorithms. For more details about these algorithms, we refer the reader to the survey of Mooney and Roddick \cite{Mooney2013}.

Many algorithms extend the principles of these algorithms to extract alternative forms of sequential patterns. Constraints and condensed patterns are among the most studied alternative patterns due to their relevance to a wide range of applications or to their concise representation of frequent patterns. 
Integrating constraints in sequential pattern mining is often limited to the use of anti-monotonic temporal constraints such as \texttt{maxgap} constraints. When constraints are not anti-monotonic, the previous pruning technique cannot be applied and the computation may require an exhaustive search, which is not reasonable. The usual technique consists in defining an anti-monotonic upper-bound of the measure such that a large part of the search space can be prune (\eg high occupancy patterns \cite{Zhang2015}). The tighter the upper-bound is, the better the computing performances are. However, any new type of constraint requires a long effort before being integrated in an efficient algorithm. Integrating flexible and generic constraints in a pattern mining algorithm remains a challenge.

\espace

The design of a generic framework for data mining is not a new problem. It has been especially studied within the field of inductive databases as proposed by Imielinski and Mannila \cite{Imielinski1996}.
In an inductive database, knowledge discovery is viewed as a querying process. The idea is that queries would return patterns and models. This framework is based on a parallel between database and data mining theory and has as ultimate goal the discovery of a relational algebra for supporting data mining.

In the specific field of pattern mining, designing such query languages has recently attracted interest in the literature \cite{DeRaedt15,guns2015miningzinc,negrevergne2013dominance,bonchi2006conquest,Boulicault2005,Vautier07}. For instance, Vautier et \al \cite{Vautier07} proposed a framework which is based on an algebraic specification of pattern mining operators. Bonchi et \al  \cite{bonchi2006conquest} proposed the Conquest system which is an algorithmic framework that accepts constraints with different properties (anti-monotonic, convertible, loose anti-monotonic, etc.). Boulicault and Jeudy \cite{Boulicault2005} survey the field of constraint-based data mining. Negrevergne et \al \cite{negrevergne2013dominance} recently proposed an algebra for programming pattern mining problems. This algebra allows for the generic combination of constraints on individual patterns with dominance relations between patterns.

More recently, the declarative approaches have shown a strong potential to be relevant frameworks for implementing the principles of inductive databases \cite{DeRaedt15}. Many data mining problems can be formalized as combinatorial problems in a declarative way. For instance, tasks such as the discovery of patterns in data, or finding clusters of similar examples in data \cite{DaoDV15}, often require constraints to be satisfied and require solutions that are optimal with respect to a given scoring function. 
The aim of these declarative approaches is to obtain a declarative constraint-based language even at the cost of degraded runtime performance compared to a specialized algorithm.
Three types of state-of-the-art solvers have been used: SAT solvers \cite{coquery2012sat}, CP solvers \cite{guns2015miningzinc} and ASP solvers \cite{jarvisalo2011itemset}.

MiningZinc \cite{guns2015miningzinc} is a CP-based approach providing a specific language built upon MiniZinc, a medium-level constraint modelling language \cite{nethercote2007minizinc}. 
A similar declarative language has been proposed by Bruynooghe et \al \cite{bruynooghe2015} using the IDP3 system. IDP3 is a Knowledge Base System (KBS) that intends to offer the user a range of inference methods and to make use of different state of the art technologies including SAT, SAT Modulo Theories, Constraint Programming and various technologies from Logic Programming. One example of application of their system concerns the problem of learning a minimal automaton consistent with a given set of strings.
In ASP, J\"arvisalo \cite{jarvisalo2011itemset} has proposed the first attempt of encoding pattern mining in ASP. J\"arvisalo addressed this problem as a new challenge for the ASP solver, but he did not highlight the potential benefit of this approach to improve the expressiveness of pattern mining tools. Nonetheless, the first order expressions of ASP encodings can easily be understood by users without higher abstracted languages.
Following Guns et \al's proposal \cite{guns2011CPmining}, J\"arvisalo designed an ASP program to extract frequent itemsets in a transaction database. A major feature of J\"arvisalo's proposal is that each answer set (AS) contains only one frequent itemset associated with the identifiers of the transactions where it occurs.
To the best of our knowledge, there is no comprehensive language provided for SAT-based data mining approaches.

\espace

All these approaches were conducted on itemset mining in transaction databases, which is much simpler than sequential pattern mining in a sequence database.
Some recent works have proposed to explore declarative programming for sequential pattern mining. In fact, dealing with expressive constraints is especially interesting for sequential pattern mining. The range of constraints on sequential patterns is wider than on itemsets and are meaningful for various concrete data analysis issues.

Negrevergne et Guns \cite{Negrevergne15} proposed the \cpsm\ approach which can be considered as the state of the art of declarative sequential pattern mining. Their contribution is twofold: i) the first declarative encodings of the standard sequential pattern mining task, ii) an efficient CP-based approach based on dedicated propagators that remains compatible with sequential pattern constraints. By combining efficiency and declarativity, \cpsm\ is a proof of concept that a declarative approach can be efficient to solve pattern mining tasks.

M\'etivier et \al \cite{Metivier2013} have developed a constraint programming method for mining sequential patterns with constraints in a sequence database. The constraints are based on \textit{amongst} and \textit{regular expression} constraints and expressed by automata.
Coquery et \al \cite{coquery2012sat} have proposed a SAT based approach for sequential pattern mining. The patterns are of the form $ab?c$ and an occurrence corresponds to an exact substring (without gap) with joker (the character $?$ replaces exactly one item different from $b$ and $c$).
Coletta and Negrevergne \cite{coletta2016} have proposed a purely boolean SAT formulation of sequential pattern mining (including closed and maximal patterns) that can be easily extended with additional constraints. 

ASP has also been used for sequential pattern mining \cite{Guyet_IJCAI2016,Guyet_IAF2014}.
Gebser et \al \cite{Guyet_IJCAI2016} have proposed, firstly, an efficient encoding for sequential pattern mining. Secondly, they have proposed to use the \asprin\ system for the management of pattern set constraints using preferences.
In \cite{Guyet_IAF2014}, the mining task is the extraction of serial episodes in a unique long sequence of itemsets where occurrences are the minimal occurrences with constraints.
Counting the number of occurrences of a pattern, or of a set of patterns, in a long sequence introduces additional complexity compared to mining sequential patterns from a sequence database since two pattern occurrences can overlap. The main contribution is a method for enumerating pattern occurrences that ensures the anti-monotonicity property.

\section{Experiments}\label{sec:experiments}
Having demonstrated that modelling in ASP is powerful yet simple, it is now interesting to examine the computational behavior of ASP-based encodings.

The first experiments compare the performance, in runtime and memory requirements, of the various ASP programs presented before. The objective is to better understand the advantages and drawbacks of each encoding. The questions we would like to answer are: which of the two embedding strategies is the best? 
does the encoding improvement really reduce computing resources needs? 
what is the behaviour of our encoding with added pattern constraints?

Next, we compare our results with the CP-based ones of \cpsm\ \cite{Negrevergne15}. \cpsm\ constitutes a natural reference since it aims at solving a mining task similar to the present one and since \cpsm\ adopts a semi-declarative approach, in particular, occurrence search is performed by a dedicated constraint propagator.

\espace

In all presented experiments, we use the version 4.5 of \clingo\footnote{\url{http://potassco.sourceforge.net/}}, with default solving parameters.
For benchmarking on synthetic data, the ASP programs were run on a computing server with 8Go RAM without using the multi-threading mode of \texttt{clingo}. Multi-threading reduces the mean runtime but introduces variance due to the random allocation of tasks. Such variance is inconvenient for interpreting results with repeated executions.
For real datasets, we used the multi-threading mode with 4 threads and 20Go shared RAM. This large amount of memory is required for large datasets.

\subsection{Encodings comparisons on synthetic datasets}
The first experiments were conducted on synthetic databases to control the most important features of data. It allows for an easier and more reliable analysis of time and memory requirements with respect to these parameters.
We designed a sequential database simulator to generate datasets with controlled characteristics. The generator\footnote{The generator and databases used in our experiments are available at \url{https://sites.google.com/site/aspseqmining}.} is based on a ``retro-engineering'' process: 
1) a set of random patterns is generated, 
2) occurrences of patterns are assigned to a given percentage of database sequences, and 
3) each sequence of items is randomly generated according to the patterns it must contain and a mean length.

The parameters of the generator and their default values are sum up in Table \ref{tab:generator}. Default values are those used when not explicitly specified.

The task to be solved is the extraction of the complete set of frequent patterns (see Sect. \ref{sec:spm}). It should be noted that every encoding extracts exactly the same set of patterns. Resource requirements are thus fairly comparable.
The computation runtime is the time needed to extract all the patterns. It includes both grounding and solving of the ASP programs using the quiet \clingo\ mode (no printed output).
The memory consumption is evaluated from the size of the grounded program, \ie the number of grounded atoms and rules. This approximation is accurate to compare ASP encodings. The solving process may require additional memory. This memory requirement is negligible compared to grounding.

\begin{table}[tb]
\center
\caption{Sequence generator parameters.}\label{tab:generator}
\begin{tabular}{@{\;}llp{.7\textwidth}@{\;}}
  \hline
Parameter & Default value & Description \\\hline
$D$ & 500 & number of sequences in the database\\
$l$ & 20 & sequence mean length (sequence length follows a normal law)\\
$n$ & 20 & number of different patterns\\
$lp$ & 5 & pattern mean length\\
$th_{D}$ & 10\% & minimum number of occurrences generated for each pattern\\
$k$ & 50 & alphabet size. The distribution of item occurrences follows a normal law ($\mu=.5$ and $\sigma=0.05$). Some items occur more often than others.\\\hline
\end{tabular}
\end{table}

\begin{figure}[tbp]
\centering
\includegraphics[width=\textwidth]{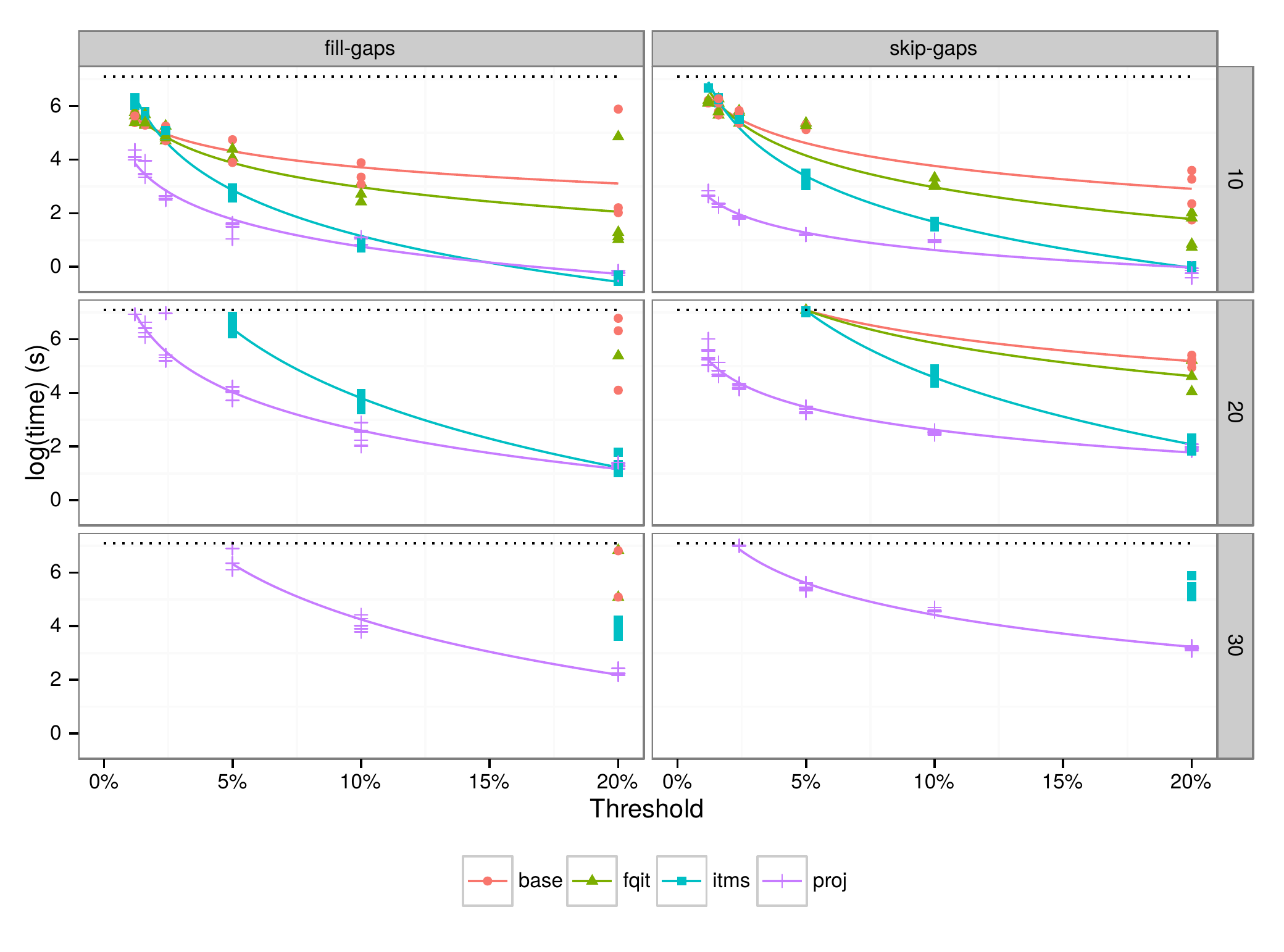}
\caption{Mean computation time for synthetic databases with sequences length from 10 to 30 (in rows).
Each curve represents an improvement to the basic encoding: none ({\sf\scriptsize base}), frequent items ({\sf\scriptsize fqit}), itemsets ({\sf\scriptsize itms}), projection ({\sf\scriptsize proj}). 
Each dot represents the mean of results on 6 datasets. 
The left-hand (resp. right-hand) column gives the results for the fill-gaps (resp. skip-gaps) strategy. The dashed horizontal line denotes the timeout of 20 minutes.}
\label{fig:expe1_cmp_time}
\end{figure}

We start with an overall comparison of the different encodings and their refinements with respect to parameters $th_{D}$ and $l$. Fig. \ref{fig:expe1_cmp_time} compares the runtimes for different encodings and the two embedding strategies, fill-gaps and skip-gaps. For each setting, 6 databases with the same characteristics were generated. Figure curves show the mean rutime of the successful executions, \ie those that extract the complete set of frequent pattern within the timeout period. The timeout was set to 20 minutes.
	
The exponential growth of the runtime when the threshold decreases is a classical result in pattern mining considering that the number of patterns grows exponentially. Every approach conforms to this behaviour. In more details:
\begin{itemize}
\item the longer the sequences, the greater the runtime. Most problem instances related to databases with $l=10$ can be solved by any approach. When the mean length of sequences increases, the computation time increases also and the number of instances solved within the timeout period  decreases. This can be easily explained by the combinatorics of computing embeddings which increases with the sequence length.
\item all proposed improvements do improve runtime for high frequency thresholds on these small synthetic databases. For $f_{min}=20\%$, the curve of every proposed improvement is below the curve of the basic encoding. For high thresholds, prefix-projection and itemsets improvements are significantly better. 
Nonetheless, the lower the threshold, the lower the difference between computation times. This shows that, except for prefix-projection, the improvements are not so efficient for hard mining tasks.
\item the prefix-projection improvement is the fastest and reduces significantly the computation time (by 2 to 3 orders of magnitude).
\item the skip-gaps strategy is more efficient than the fill-gaps strategy for these small datasets. The skip-gaps strategy requires less time to extract the same set of patterns than the fill-gaps strategy, for the same encoding improvements.
\end{itemize}

We will see below that this last result does not accurately predict which strategy should be preferred for mining real datasets. Before, we analyse the memory requirements of the different encodings.

\begin{figure}[tbp]
\centering
\includegraphics[width=\textwidth]{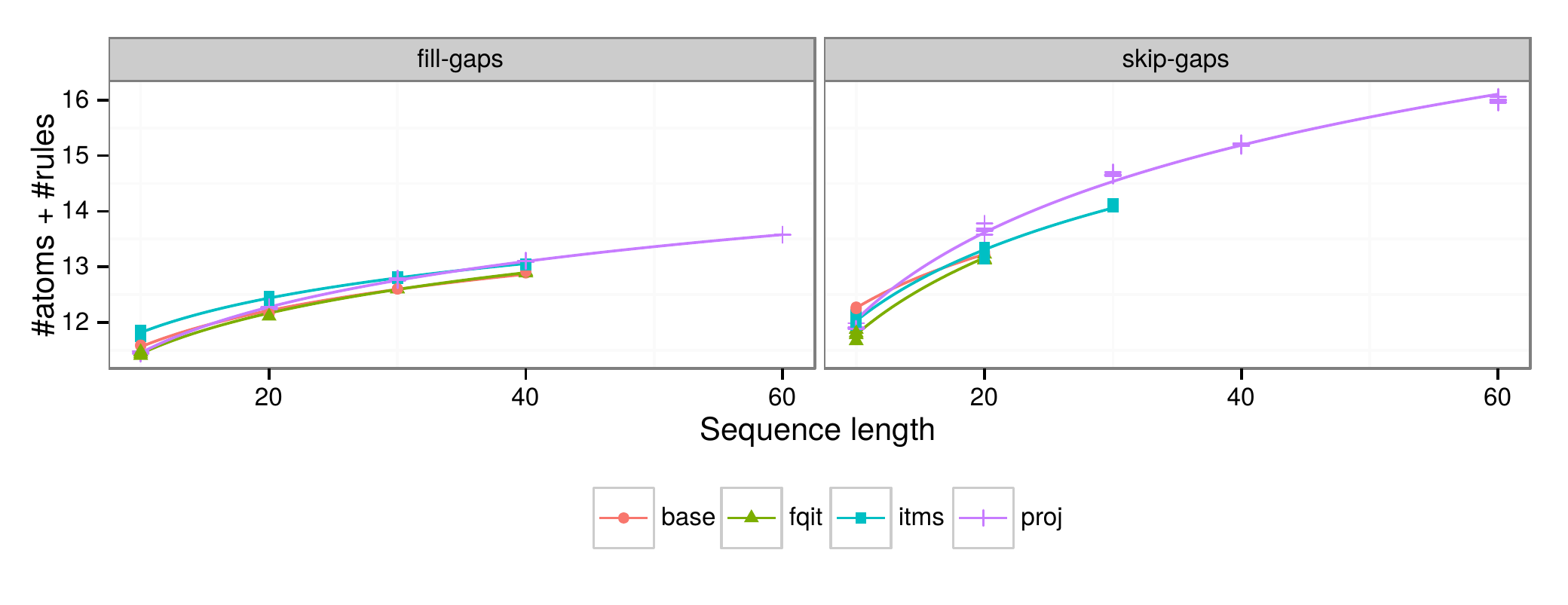}
\caption{Memory requirement with respect to sequence length. Problem size (estimated memory requirement) for synthetic databases of sequences of length from $10$ to $30$ (in rows), under frequency threshold of $20\%$.}
\label{fig:expe1_cmp_mem}
\end{figure}

\espace

We first note that the memory consumption is not related to the frequency threshold. This is a specificity of declarative pattern mining. Thus, Fig. \ref{fig:expe1_cmp_mem} compares the embedding strategies only for a unique frequency threshold $f_{min} = 20\%$.
The curves show the number of grounded atoms and rules. As it represents a tight approximation of the memory requirement, we will refer to memory in the sequel.

Unsurprisingly, the richer the encoding is, the more memory is required. But the differences are not really significant, except for the prefix-projection programs ({\sf\footnotesize proj}) which requires the highest number of atoms. 
We can see that using frequent itemsets ({\sf\footnotesize itms}) is efficient to reduce the memory requirement. This means that the grounding step was able to exploit the additional rules to avoid the creation of useless atoms and rules. Such a kind of rules is really interesting because, as the algorithmic complexity of the mining task is not high, the efficiency of the ASP program is related to his grounding size.

\begin{figure}[tbp]
\sidecaption
\centering
\includegraphics[width=0.6\textwidth]{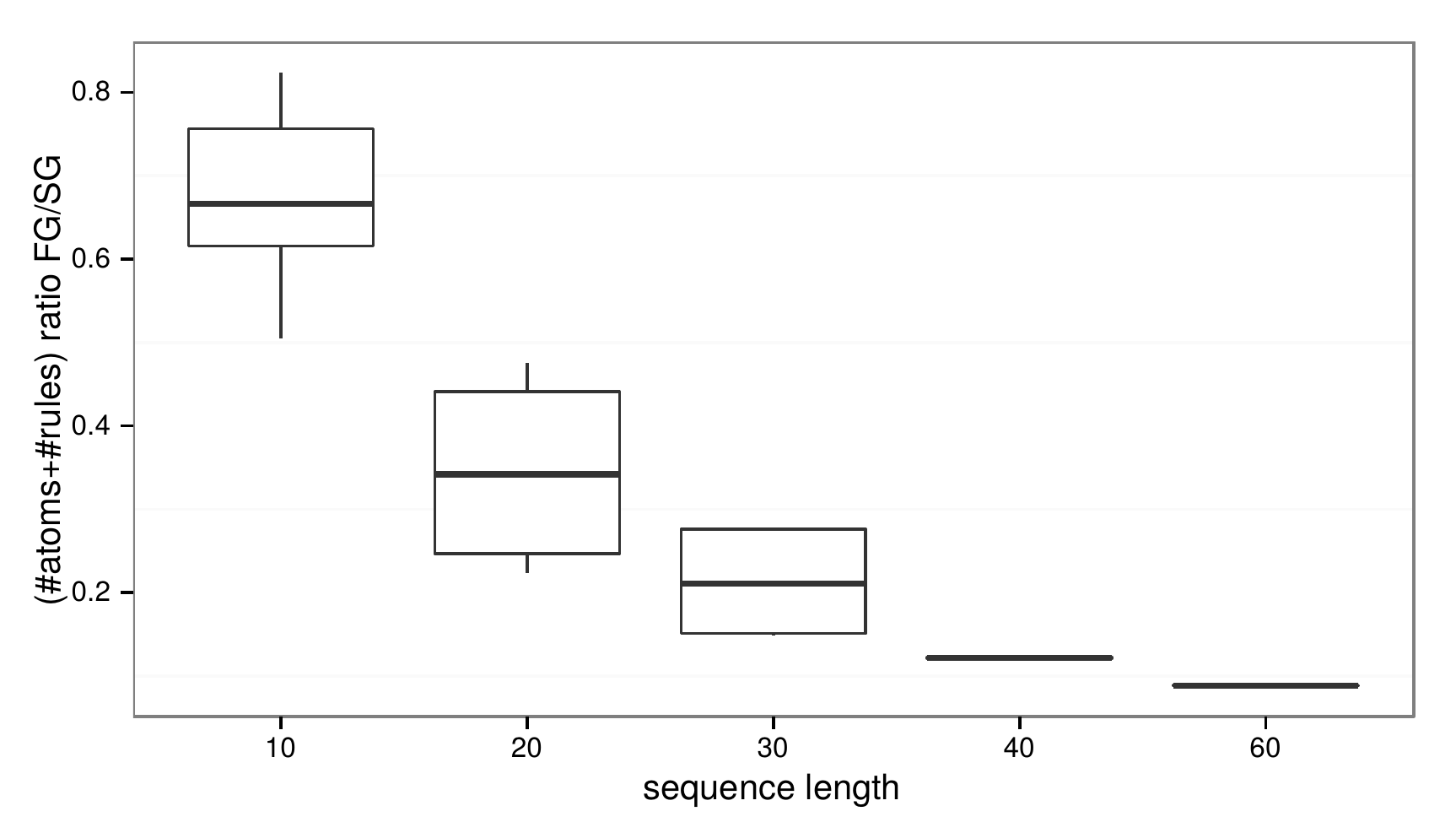}
\caption{Dispersion of ratios of memory consumption obtained for the skip-gaps strategy to those obtained for the fill-gaps strategy. Boxplots were computed for problem instances with threshold at $20\%$ and for all lengths and all encodings.}
\label{fig:expe1_fgsg_memratio}
\end{figure}

In addition, from this last point of view, we can note that the fill-gaps strategy requires several order less memory than the skip-gaps strategy.
The longer the sequences, the larger the difference. This result is illustrated by Fig. \ref{fig:expe1_fgsg_memratio}. For each problem instance, the ratio of memory usage is computed by dividing the memory required by encoding with skip-gaps strategy with the memory required by the similar encoding with the fill-gaps strategy. Fig. \ref{fig:expe1_fgsg_memratio} illustrates with boxplots the dispersion of these ratios for different sequence lengths.
Fig. \ref{fig:expe1_fgsg_memratio} clearly shows that the longer the sequences are, the more efficient the fill-gaps strategy is for memory consumption.

\espace

To end this overall comparison, it is interesting to come back to runtime. The overall results of Fig. \ref{fig:expe1_cmp_time} show that the skip-gaps strategy seems better, but considering that the fill-gaps strategy requires less memory, it is interesting to analyse the evolution of computation time with respect to database size.

\begin{figure}[tbp]
\sidecaption
\centering
\includegraphics[width=0.6\textwidth]{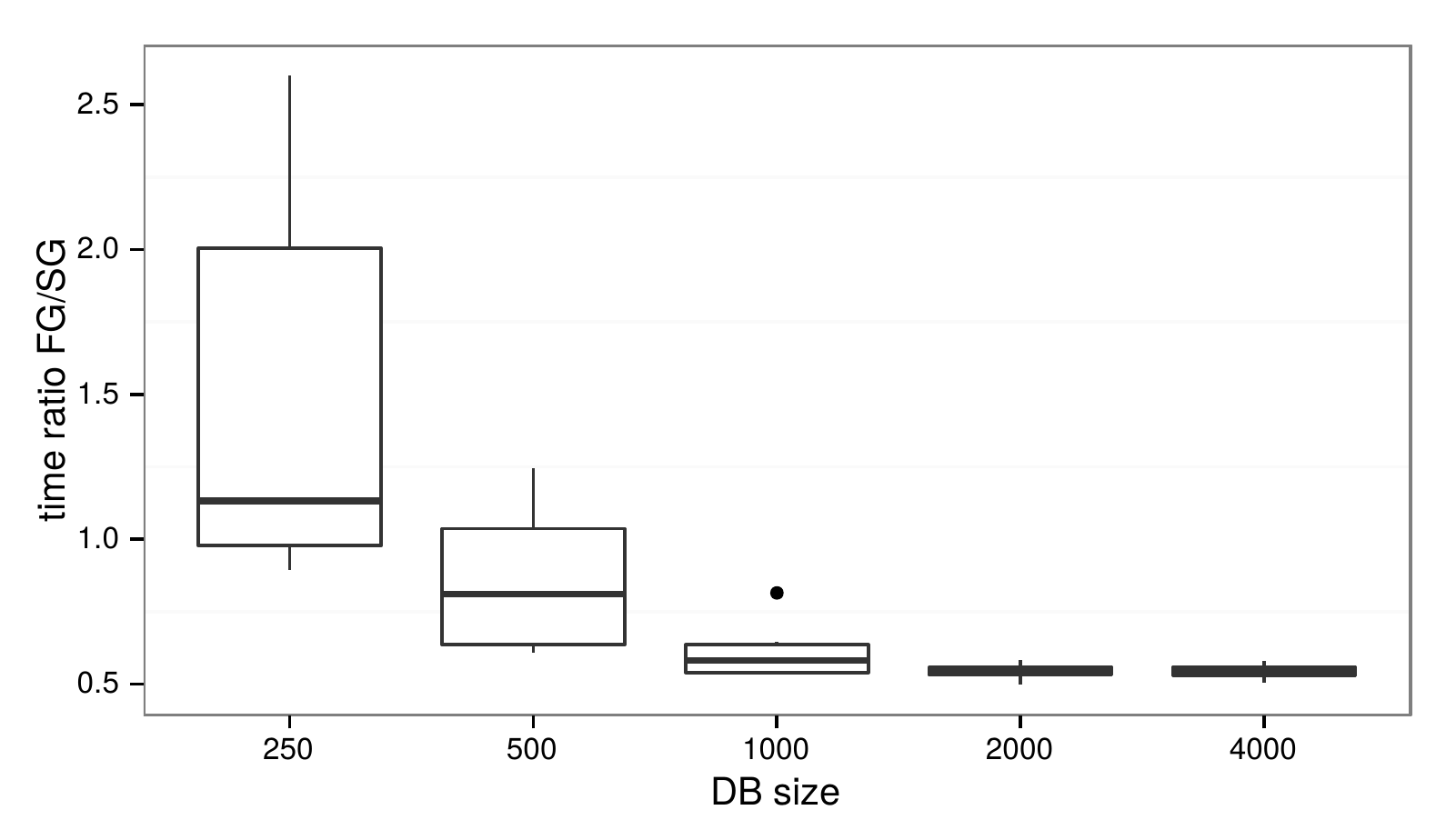}
\caption{Dispersion of ratios of runtime of the skip-gaps strategy to the runtime of the fill-gaps strategy.}
\label{fig:expe3_fgsg_dbsize}
\end{figure}

Fig. \ref{fig:expe3_fgsg_dbsize} illustrates the ratio of runtimes in both strategies when the database size increases. The support threshold, $f_{min}$, is fixed to $10\%$ and the sequence mean length to $20$. We used the prefix-projection encoding for this experiment.
Similarly to the previous figure, the ratios were individually computed for each pair of results (fill-gaps/skip-gaps) and the figure shows statistics about these ratio.

Fig. \ref{fig:expe3_fgsg_dbsize} shows clearly that when the database size increases, the fill-gaps strategy becomes more efficient than the skip-gaps strategy.

\espace

From these experiments, we can conclude that combining prefix-projection  with the fill-gaps strategy gives the best encoding. Thus, in the next subsection, we will compare this encoding with \cpsm.

\subsection{Real dataset analysis}
In these experiments, we analyse the proposed encodings on processing real datasets. We use the same real datasets as selected in \cite{Negrevergne15} to have a representative panel of application domains:
\begin{itemize}
\item JMLR: a natural language processing dataset; each transaction is a paper abstract from the Journal of Machine Learning Research,
\item UNIX: each transaction is a series of shell commands executed by a user during one session,
\item iPGR: each transaction is a sequence of peptides that is known to cleave in presence of a Trypsin enzyme,
\item FIFA: each transaction is a sequence of webpages visited by a user during a single session.
\end{itemize}
The dataset characteristics are sum up in Table \ref{tab:dataset}. Some of them are similar to those of simulated datasets.

\begin{table}[tb]
\centering
\caption{Dataset characteristics: alphabet size, number of sequences and items, max and mean length of sequences, dataset density.}
\label{tab:dataset}
\begin{tabular}{lcccccc}
\textbf{Dataset} & $|\mathcal{I}|$ & $|D|$ & $||D||$ & $max |T |$ & $avg |T |$ & $density$ \\ \hline
\textbf{Unix user} & 265 & 484 & 10935 & 1256 & 22.59 & 0.085 \\
\textbf{JMLR} & 3847 & 788 & 75646 &231 & 96.00 & 0.025 \\
\textbf{iPRG} & 21 & 7573 & 98163 & 13 & 12.96 & 0.617 \\
\textbf{FIFA} & 20450 & 2990 & 741092 & 100 & 36.239 & 0.012
\end{tabular}
\end{table}

\subsubsection{Comparison of frequent pattern mining with \cpsm}

\begin{figure}[t]
\centering
\includegraphics[trim=0 0 109 0, clip=true, width=.48\textwidth]{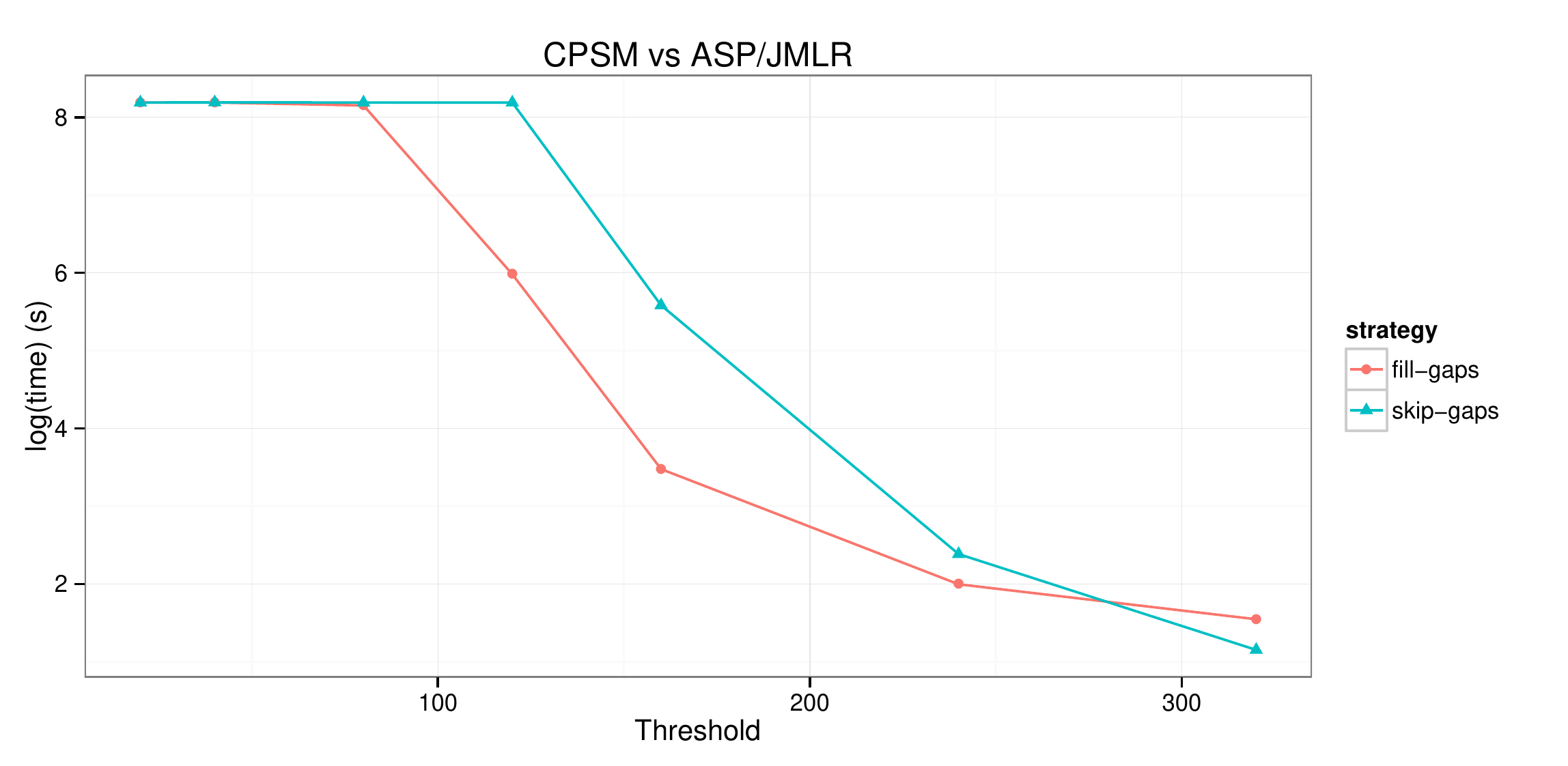}
\includegraphics[trim=0 0 109 0, clip=true, width=.48\textwidth]{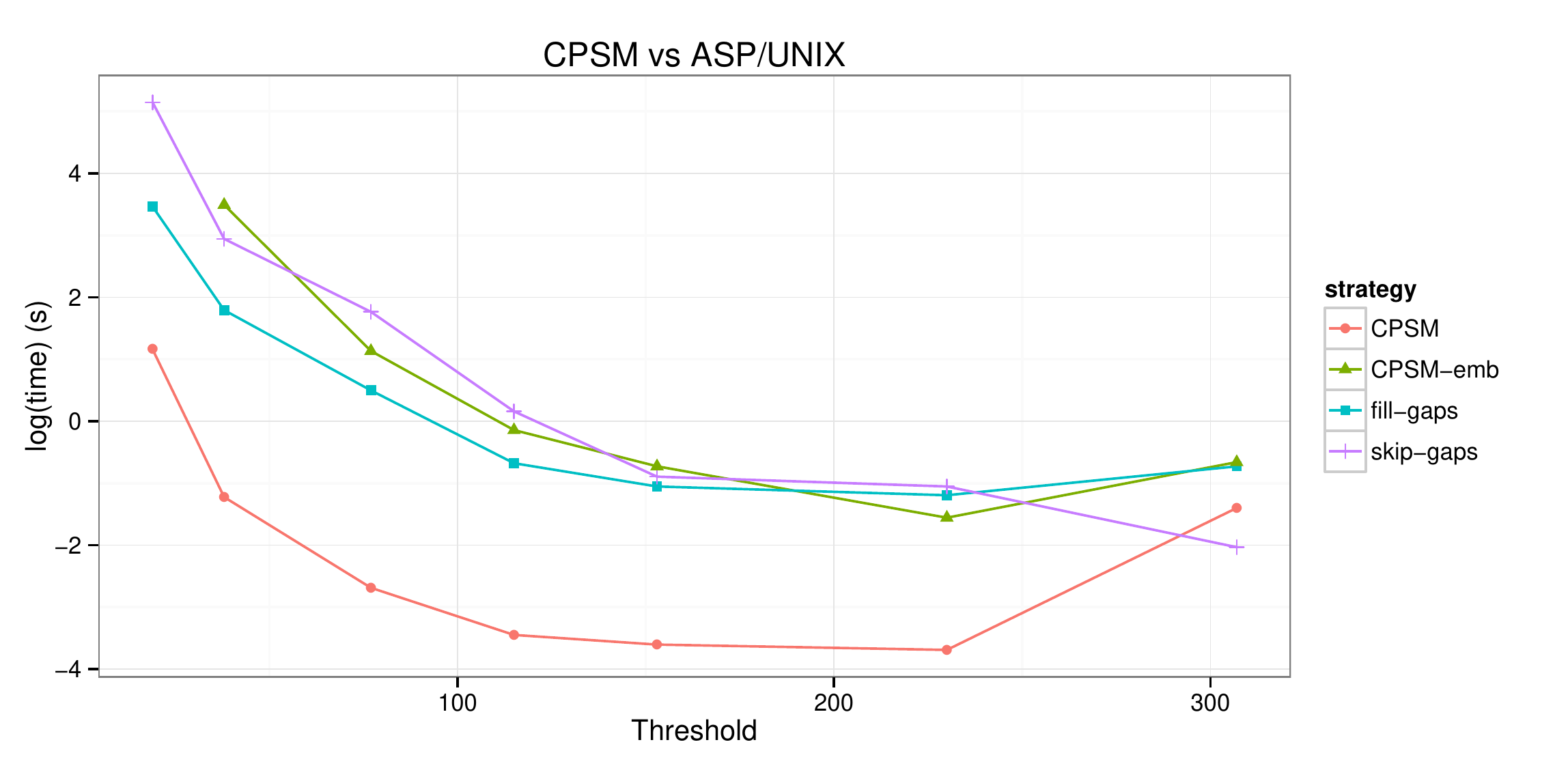}\\
\includegraphics[trim=0 0 109 0, clip=true, width=.48\textwidth]{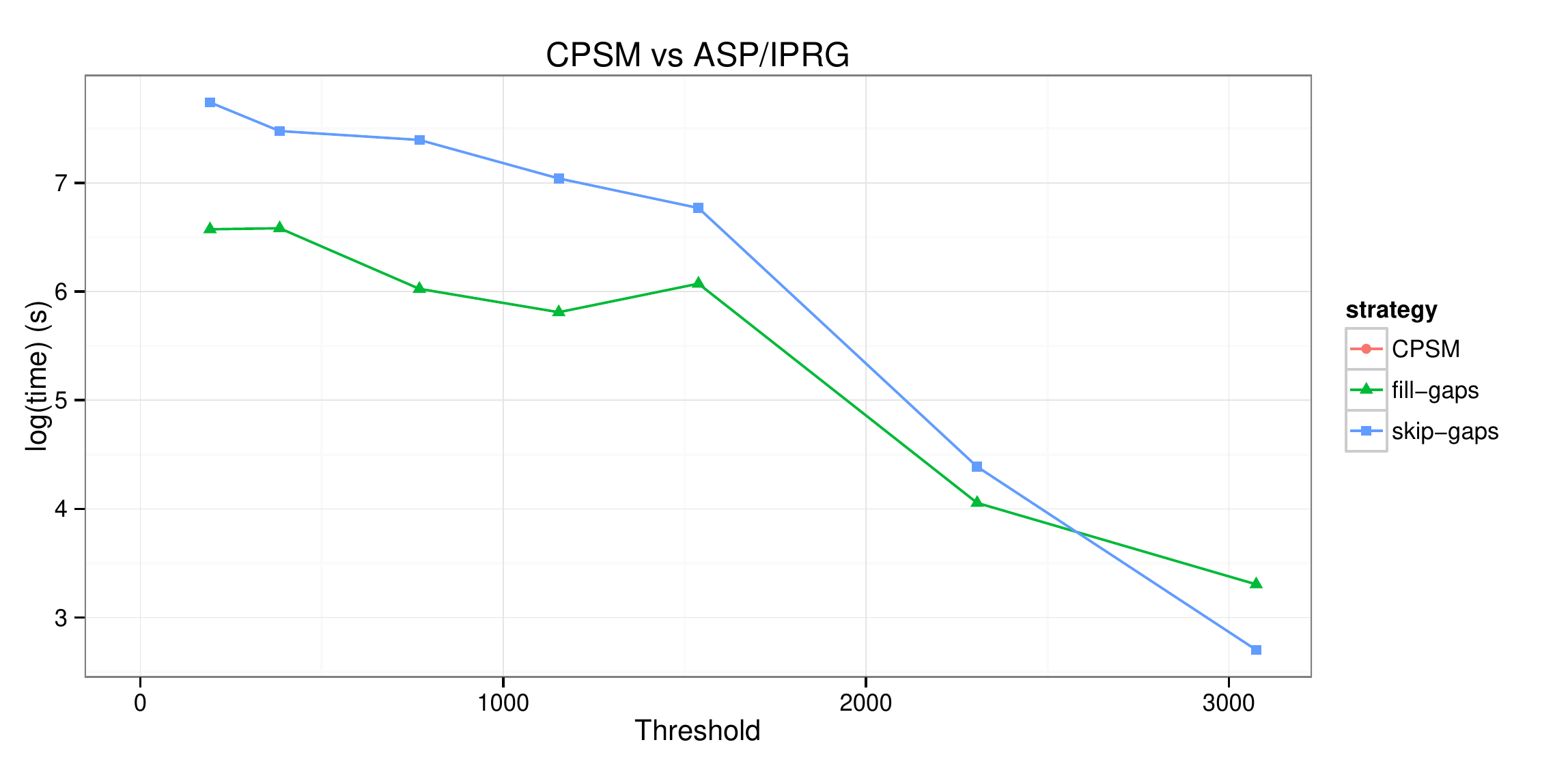}
\includegraphics[trim=0 0 109 0, clip=true, width=.48\textwidth]{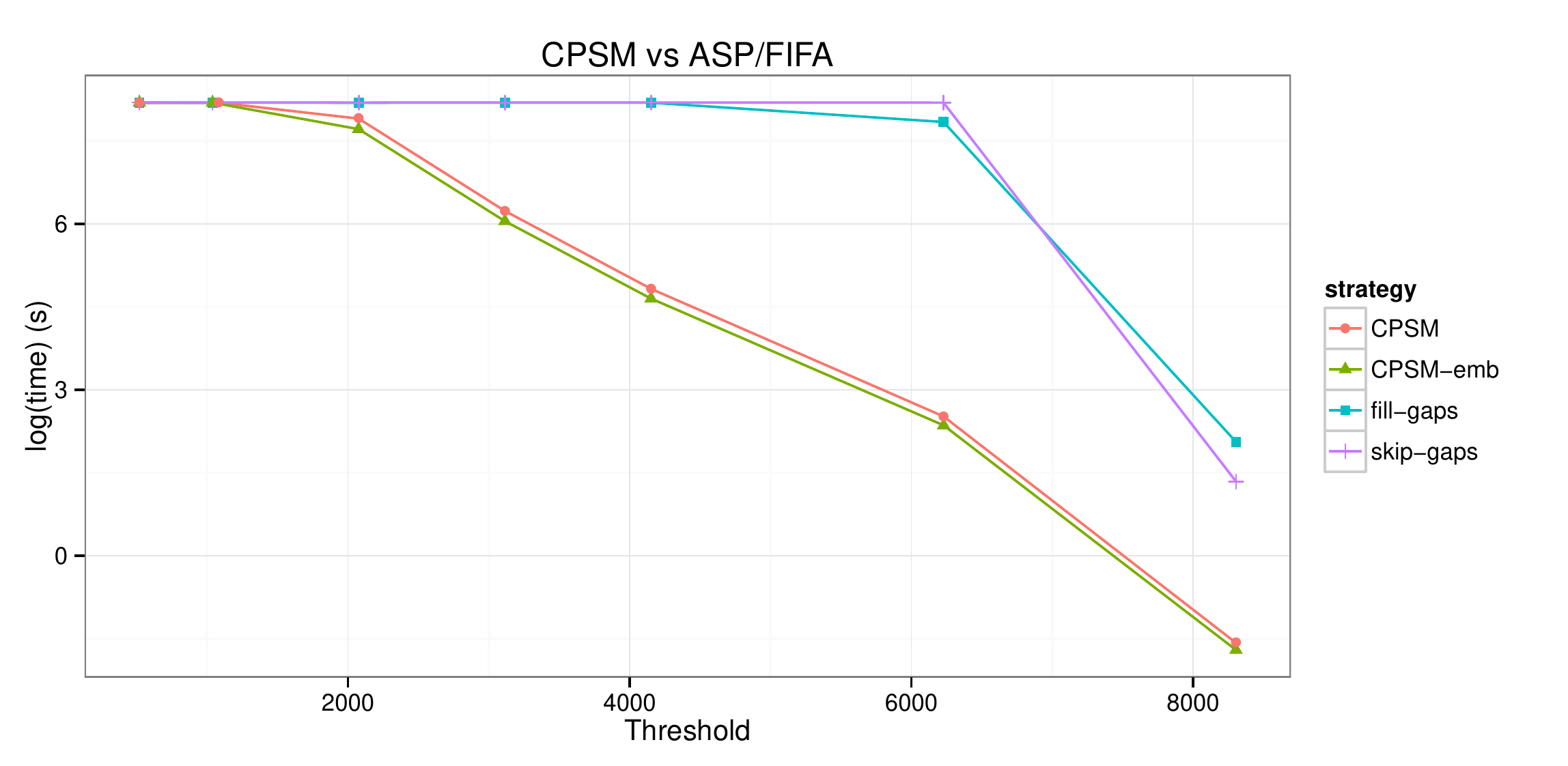}\\
\includegraphics[trim=78 70 35 70, clip=true, width=.4\textwidth]{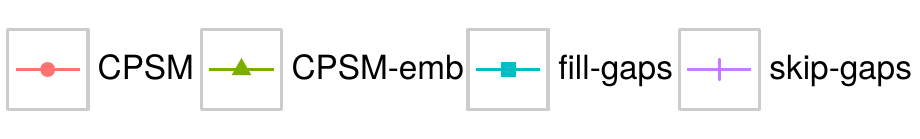}
\caption{Runtime for mining frequent patterns with four approaches: \cpsm, \cpsmemb, ASP with fill-gaps, ASP with skip-gaps.}
\label{fig:cpsm_vs_asp}
\end{figure}

Fig.~\ref{fig:cpsm_vs_asp} compares the runtimes of ASP-based sequence mining (using the ASP system \clingo) and \cpsm\ (based on the CP solver \gecode).
We ran the two versions of \cpsm. \cpsm\ makes use of global constraints to compute embeddings. This version is known to be very efficient, but it cannot cope with embedding constraints, while \cpsmemb\  does but is less efficient.
We do not compare our approach with dedicated algorithms, 
which are known to be more efficient than declarative mining approaches (see \cite{Negrevergne15} for such comparisons).
The timeout was set to 1 hour.

The results show that the runtimes obtained with \clingo\ are comparable to \cpsmemb. It is lower for IPGR, very similar for UNIX and larger for JMLR. 
These results are consistant to those presented in \cite{Guyet_IJCAI2016} for synthetic datasets. When sequences become large, the efficiency of our encoding decreases somewhat. The mean length for JMLR is $96$ while it is only $12.96$ for iPRG. 
For \cpsm\  with global constraints, the runtime-efficiency is several order of magnitude faster. 
To be fair, it should be noted that ASP approach ran with four parallel threads while \cpsmemb\ ran with no  multi-threading since it does not support it.
It should also be noted that \cpsm\ requires a lot of memory, similarly to ASP-based solving.

\subsubsection{Comparison of constrained frequent pattern mining with \cpsm}
In this section, we detail the performance on constrained pattern mining tasks. We compare our approach with \cpsmemb, which enables \emph{max-gap} and \emph{max-span} constraints. In this experiments we took the same setting as the experiments of \cite{Negrevergne15}: we add first a constraint  \emph{max-gap=2} and then we combine it with a second constraint \emph{max-span=10}. For each setting, we compute the frequent patterns with our ASP encoding and with \cpsm\ for the four datasets.

\begin{figure}[t]
\centering
\includegraphics[trim=0 0 109 0, clip=true, width=.48\textwidth]{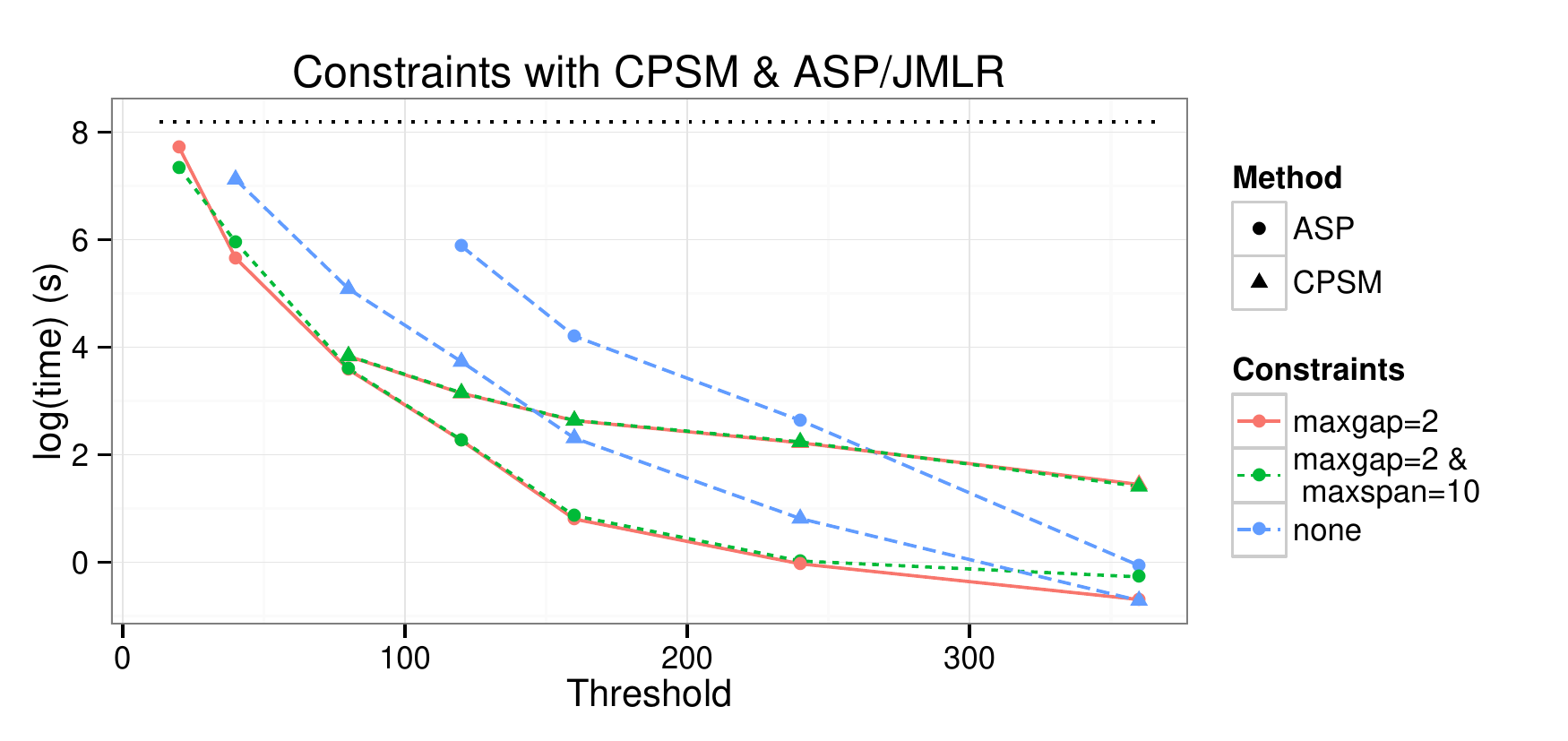}
\includegraphics[trim=0 0 109 0, clip=true, width=.48\textwidth]{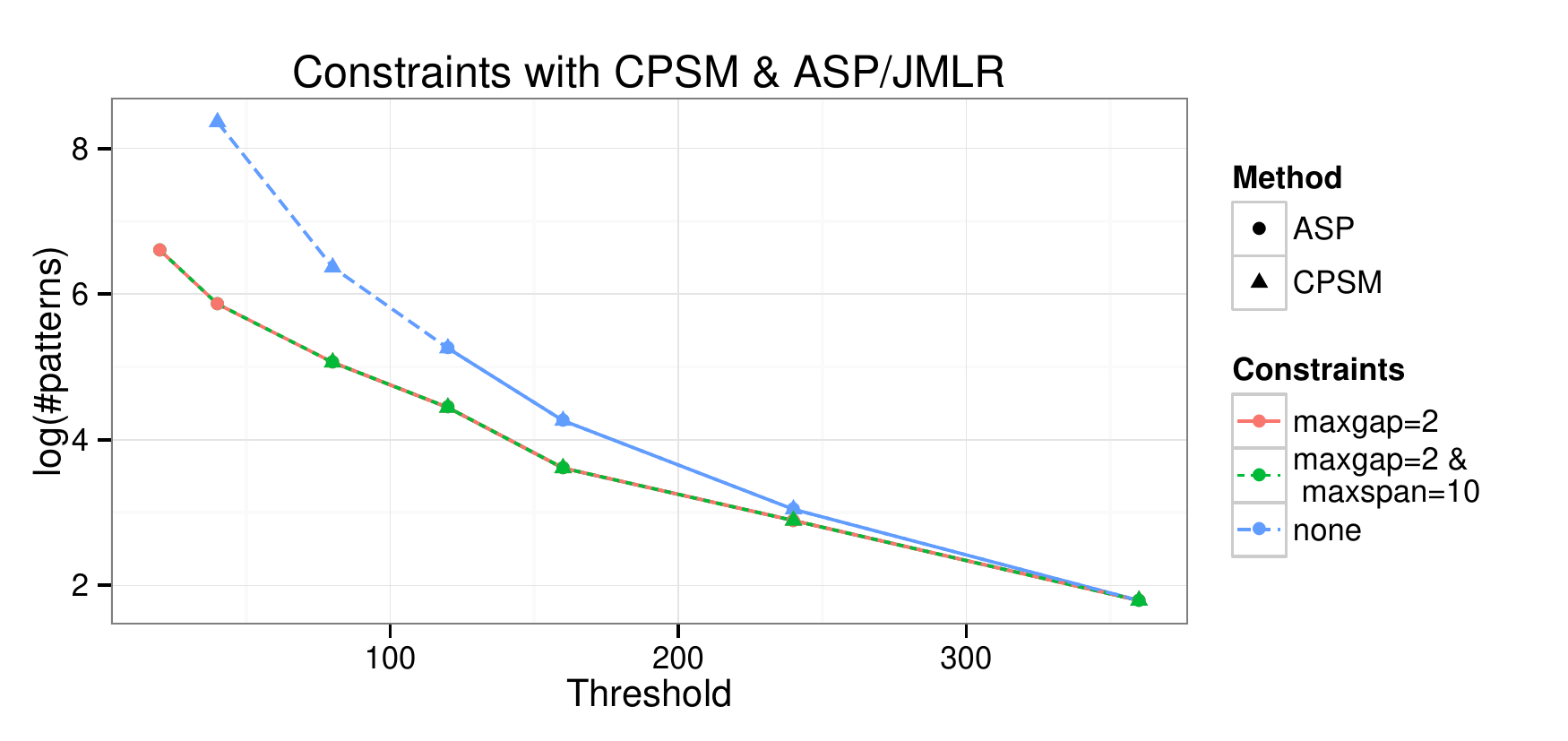}\\
\includegraphics[trim=0 0 109 0, clip=true, width=.48\textwidth]{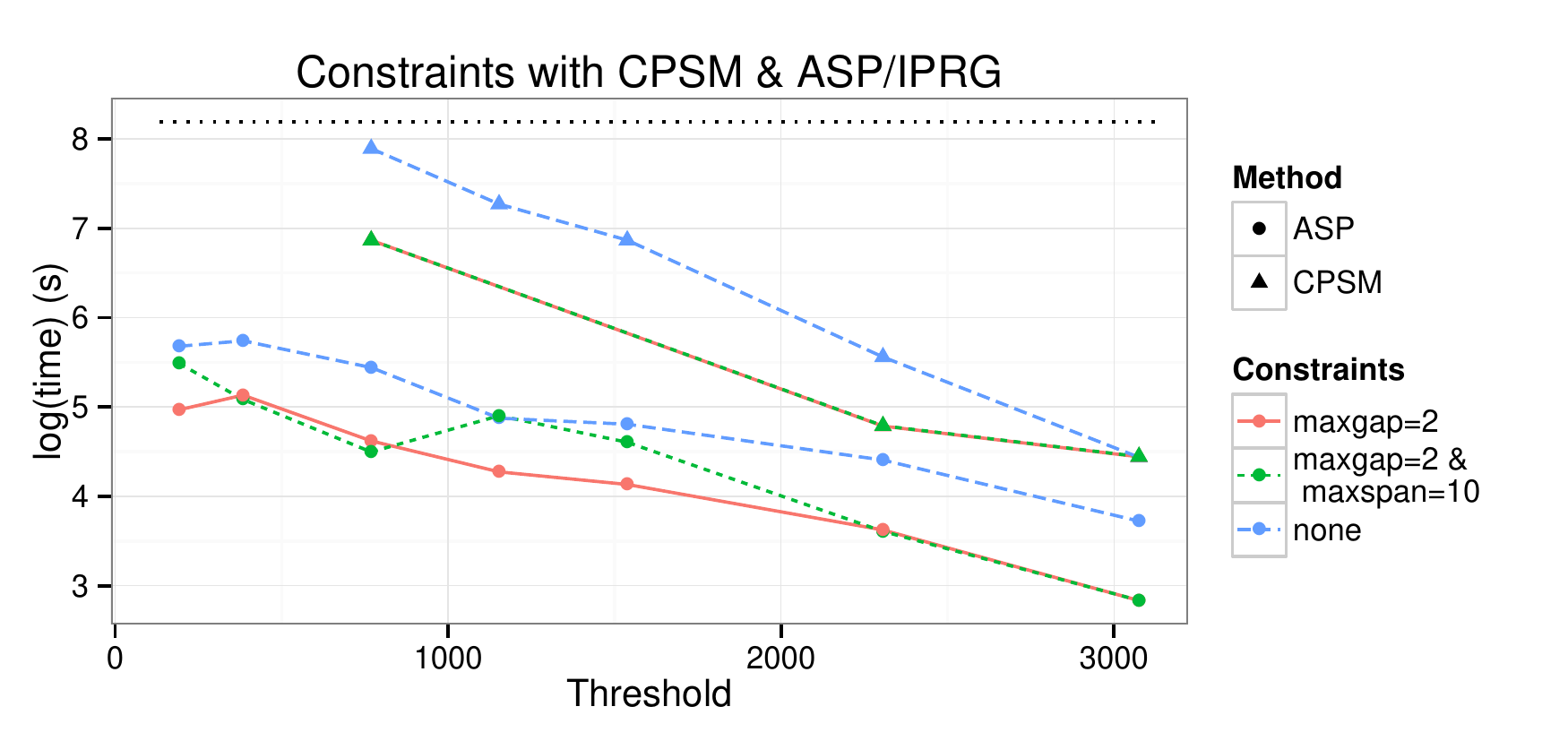}
\includegraphics[trim=0 0 109 0, clip=true, width=.48\textwidth]{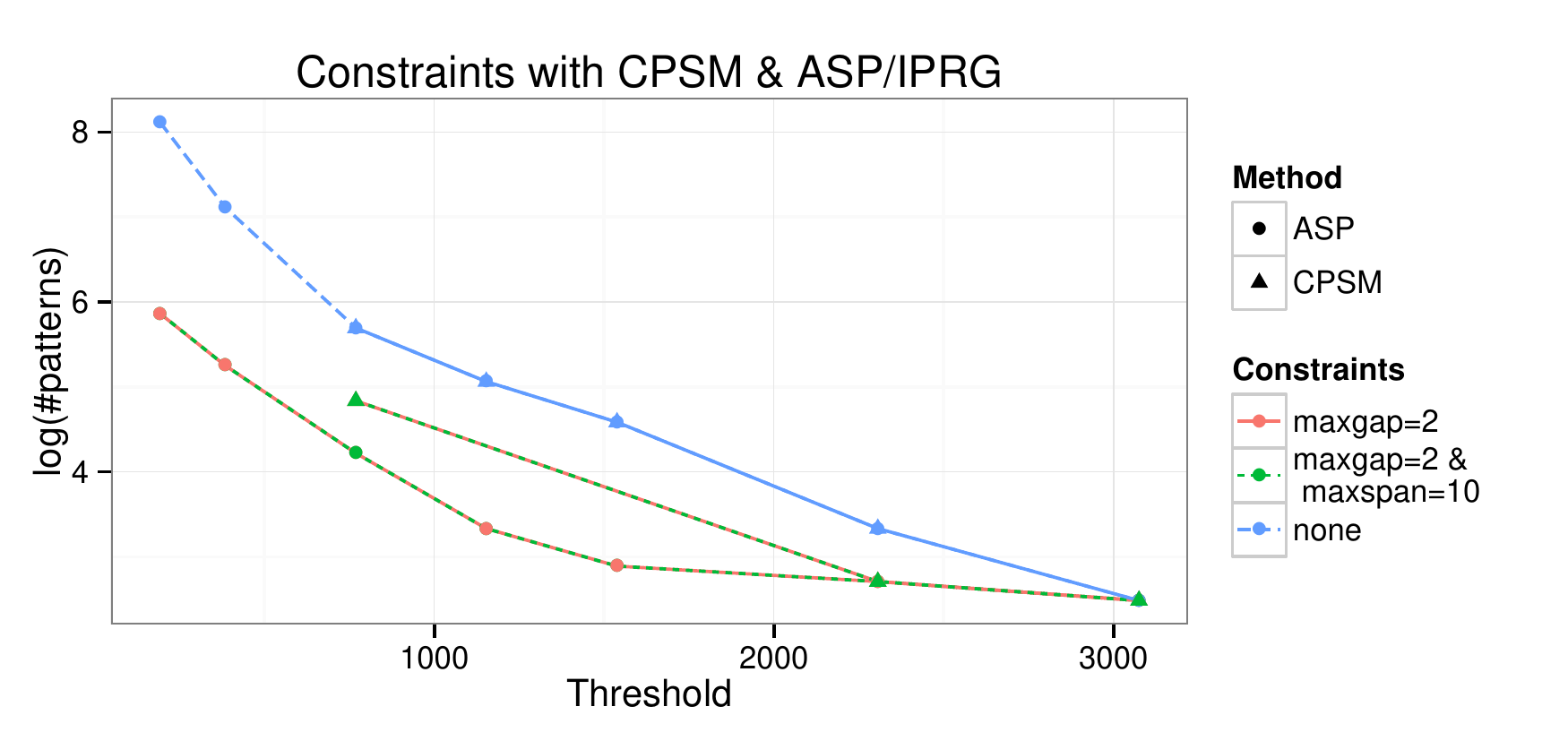}\\
\includegraphics[trim=0 0 109 0, clip=true, width=.48\textwidth]{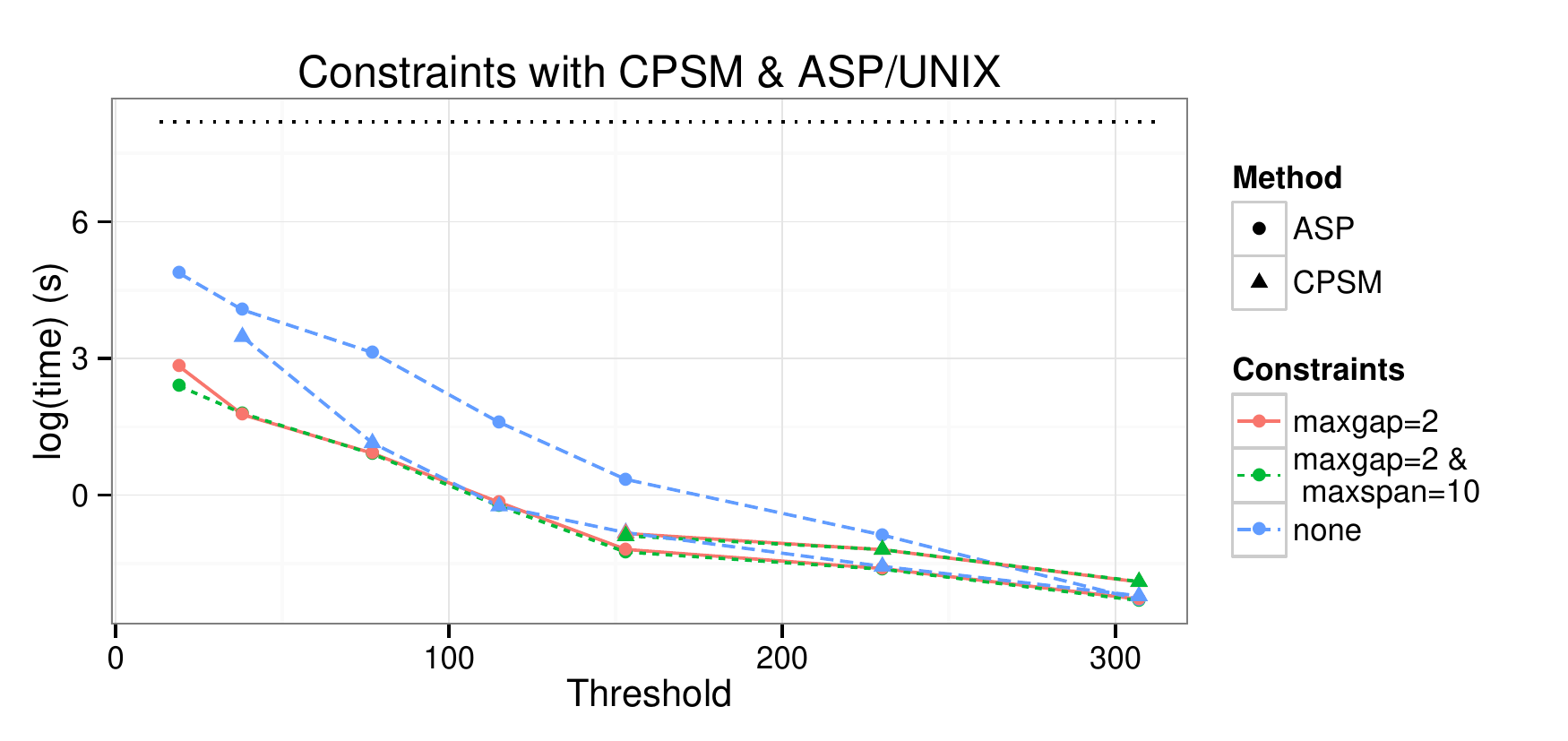}
\includegraphics[trim=0 0 109 0, clip=true, width=.48\textwidth]{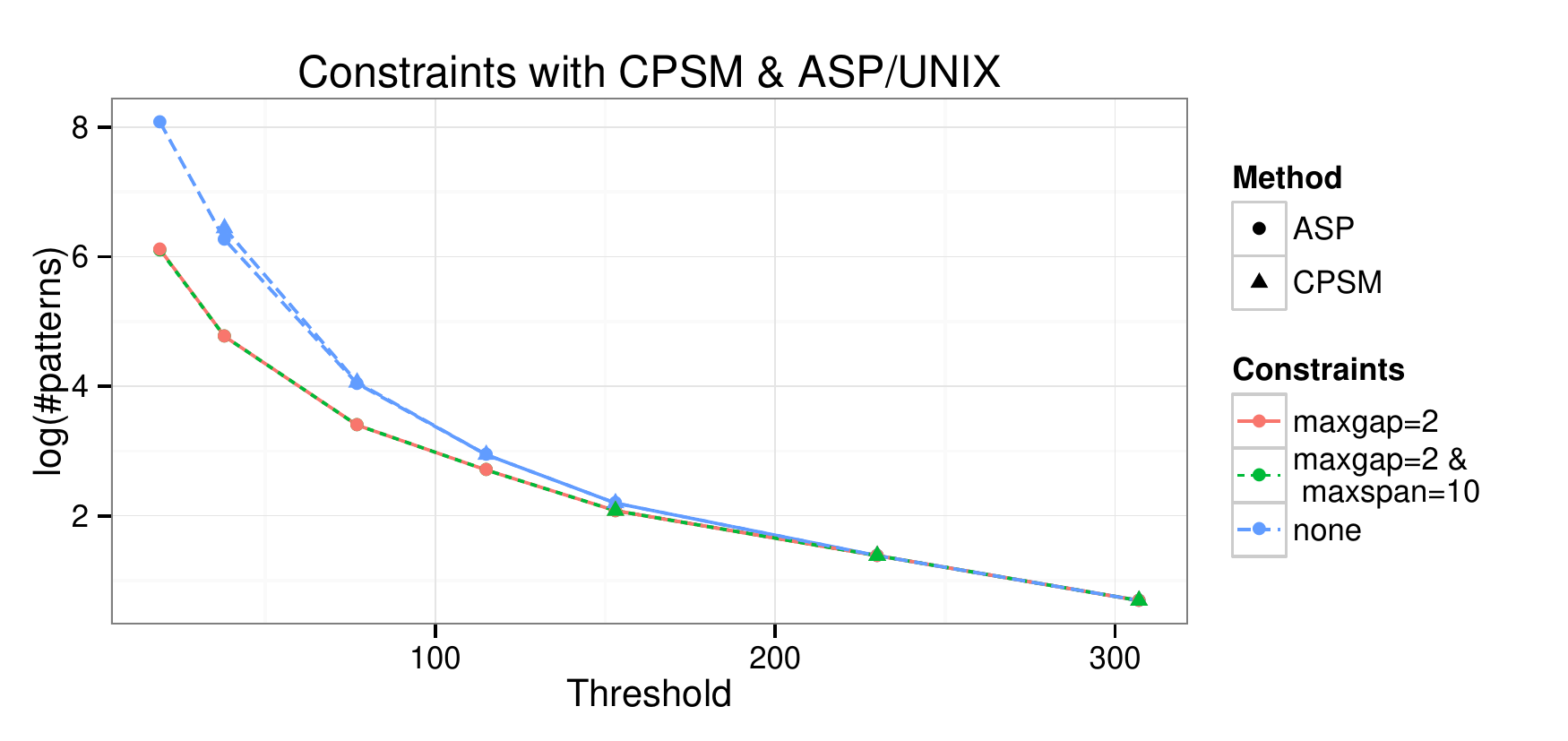}\\
\includegraphics[trim=0 0 109 0, clip=true, width=.48\textwidth]{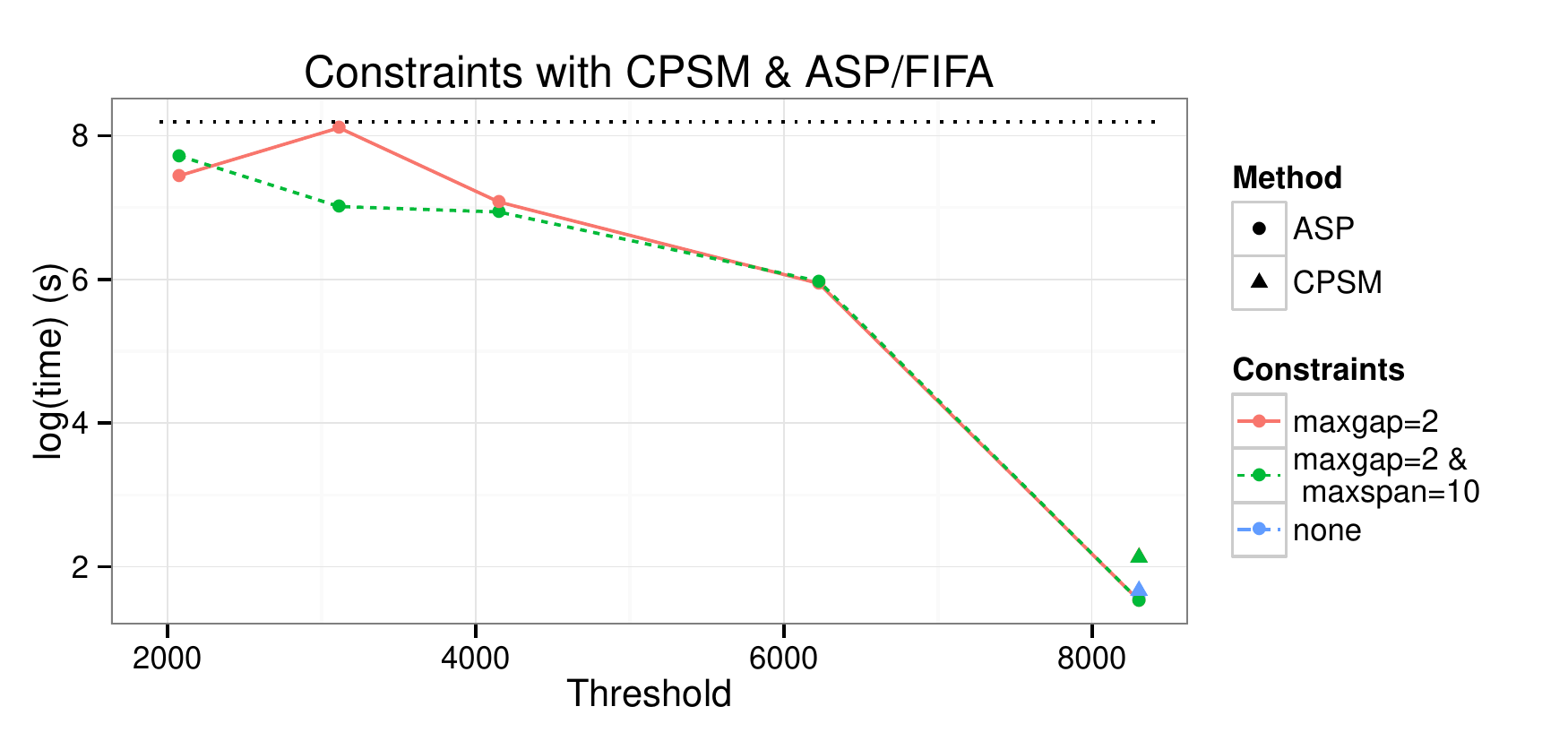}
\includegraphics[trim=0 0 109 0, clip=true, width=.48\textwidth]{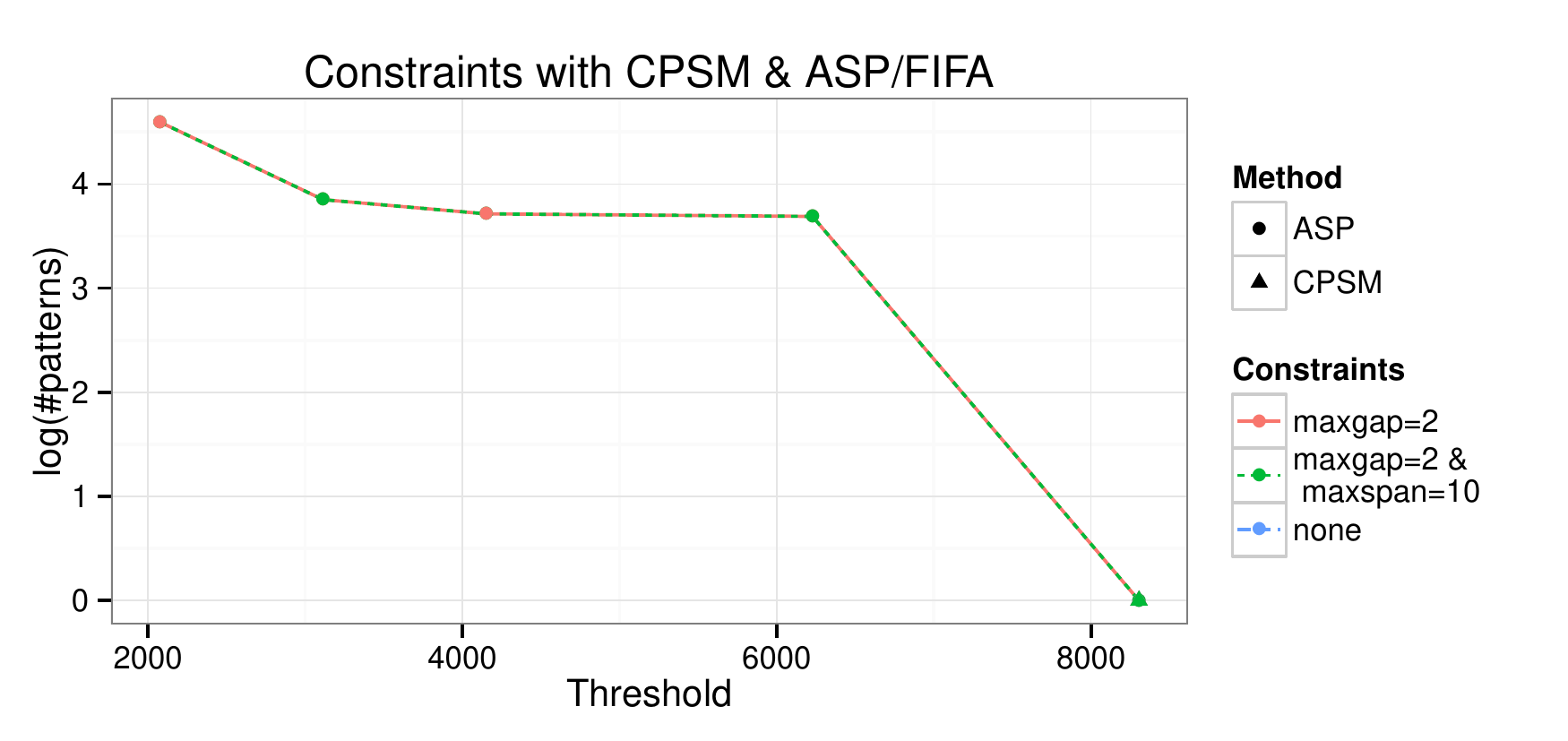}\\
\includegraphics[trim=55 58 55 58, clip=true, width=.4\textwidth]{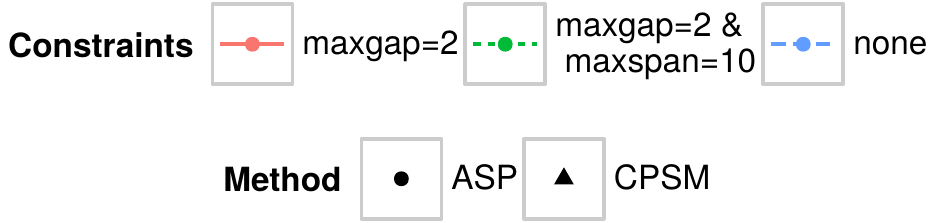}
\caption{Results for constrained sequence mining tasks with ASP vs \cpsm. On the left: runtime; on the right: number of patterns. The two figures are in log scales. From top to bottom, JMLR, IPRG, UNIX and FIFA. For each plot, the curves illustrate the results for different type of constraints (see legend). The horizontal dashed line figure out the 1 hour timeout.}
\label{fig:cpsm_vs_asp_constrainted}
\end{figure}

Fig. \ref{fig:cpsm_vs_asp_constrainted} shows the runtime and the number of patterns for each experiment. This figure illustrates results for completed searches. A first general remark is that adding constraints to ASP encodings reduces computation times. Surprisingly for \cpsm, for some thresholds the computation with some constraints requires more time than without constraints. This is the case for example for the iPRG dataset: \cpsm\ could not solve the mining problem within the timeout period for thresholds  769 and 384. 
Surprisingly, it could complete the task for lower thresholds whereas the task should be more difficult.
ASP required also more time to solve the same problem instances, but it could complete them.
Again, we can note that the mean sequence length impacts the performance of ASP encodings. \cpsm\ has lower runtime on JMLR than ASP while it is the opposite on iPRG.

The curves related to the number of patterns demonstrate that the number of extracted pattern decreases when the number of constraints increases. Since we present only the results of completed solving, \cpsm\  and ASP yield the same set of patterns.

\subsubsection{Analysis of condensed pattern extraction}
\begin{figure}[tbp]
\centering
\includegraphics[trim=0 0 109 0, clip=true, width=.32\textwidth]{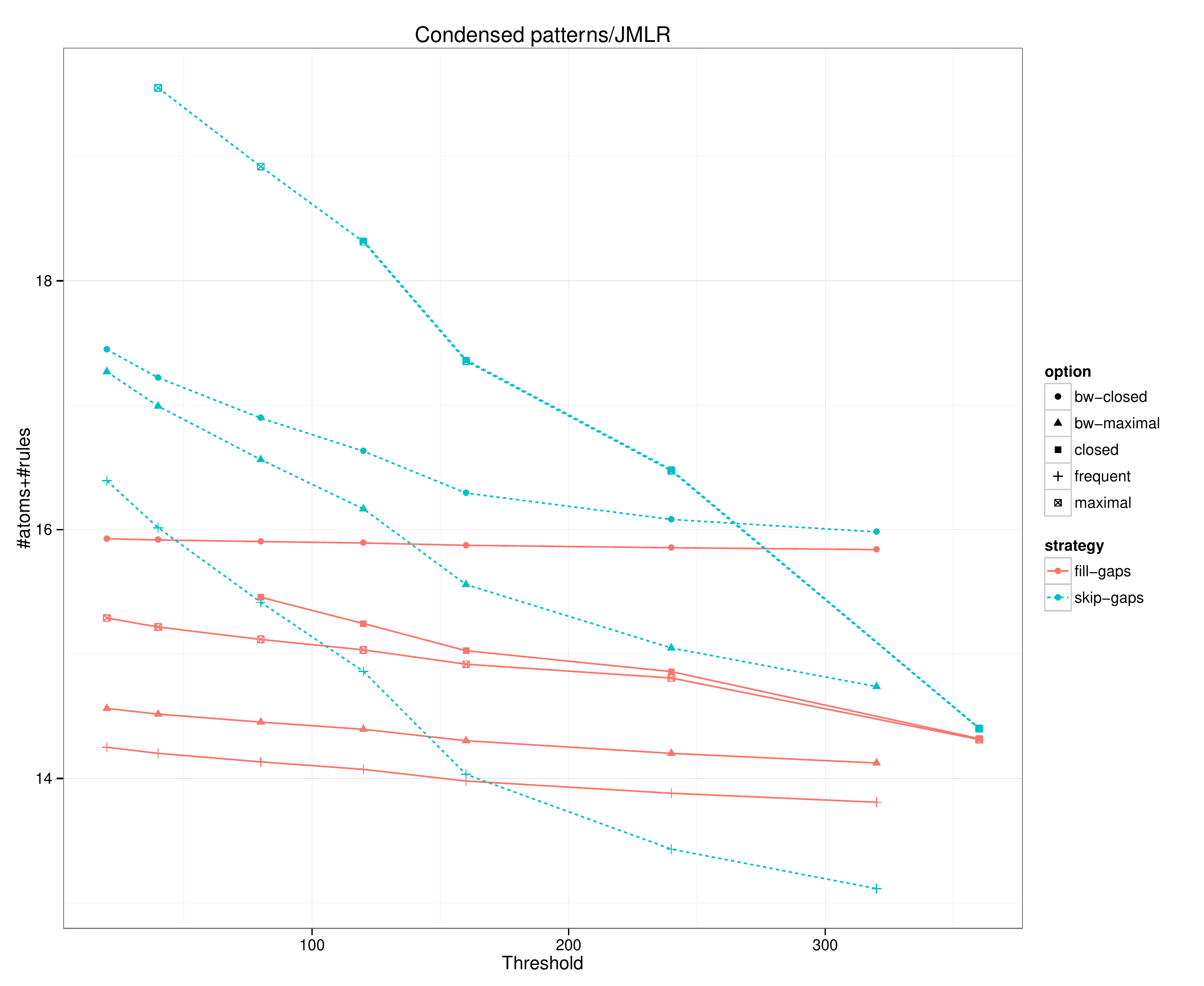}
\includegraphics[trim=0 0 109 0, clip=true, width=.32\textwidth]{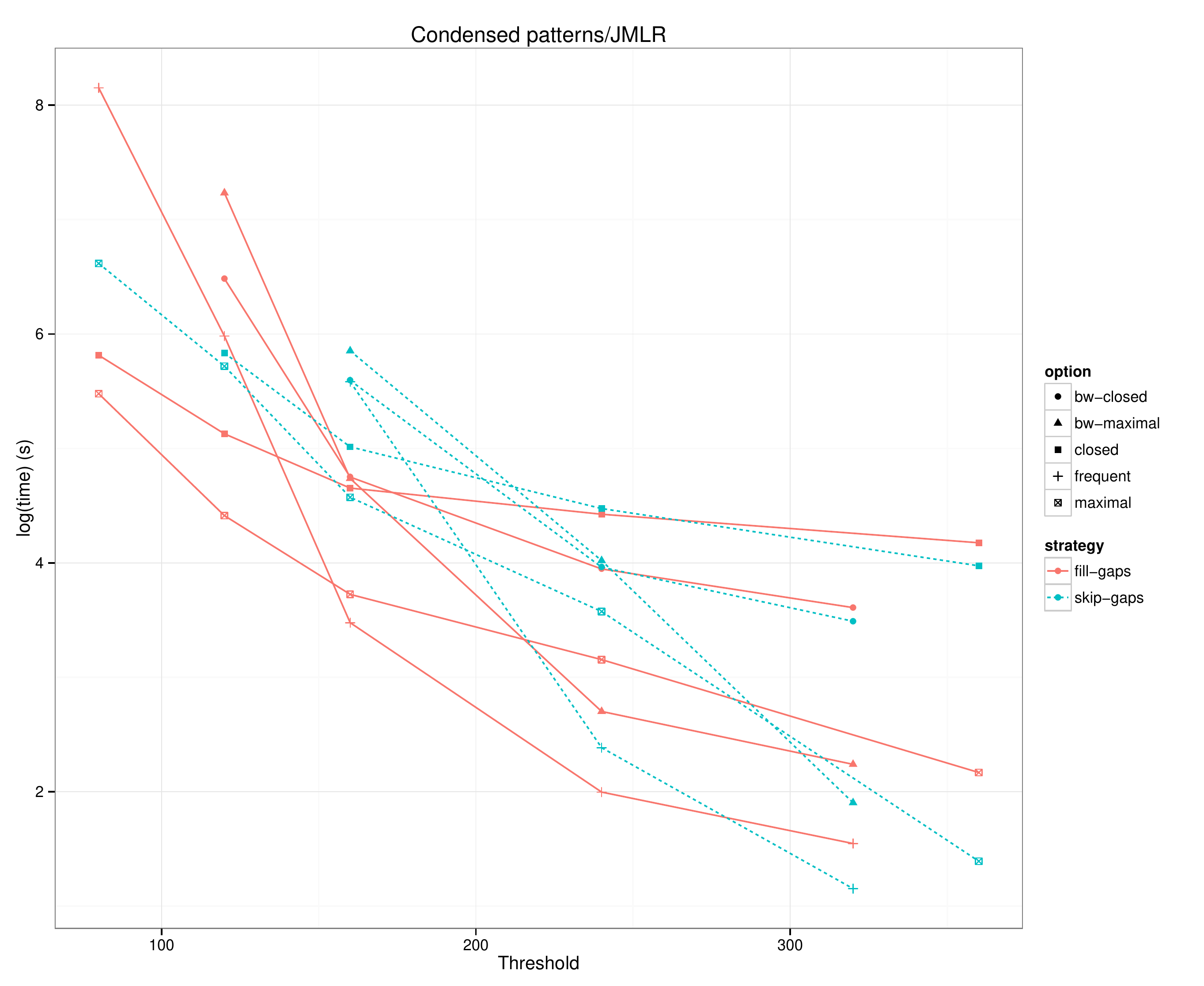}
\includegraphics[trim=0 0 109 0, clip=true, width=.32\textwidth]{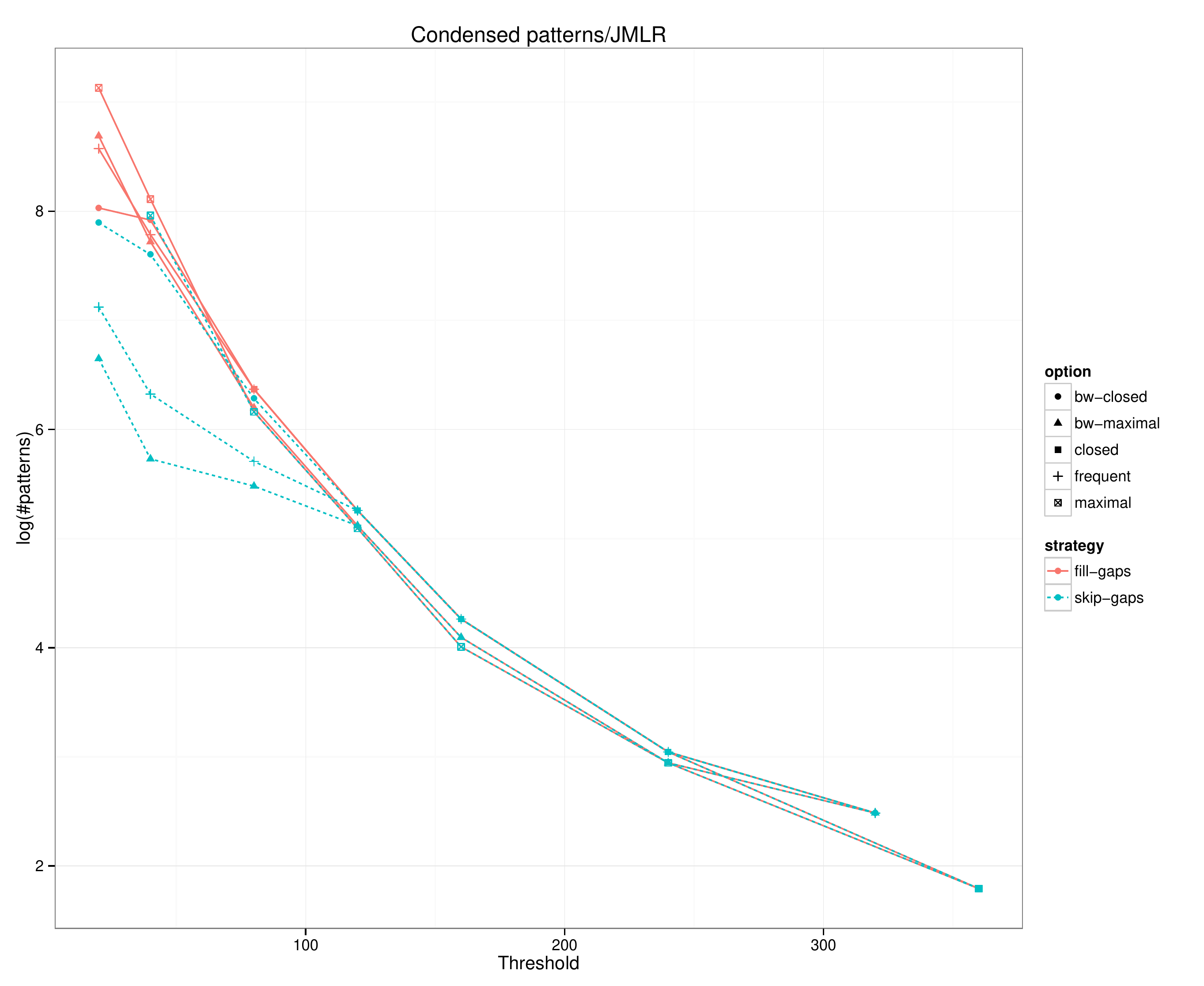}\\
\includegraphics[trim=0 0 109 0, clip=true, width=.32\textwidth]{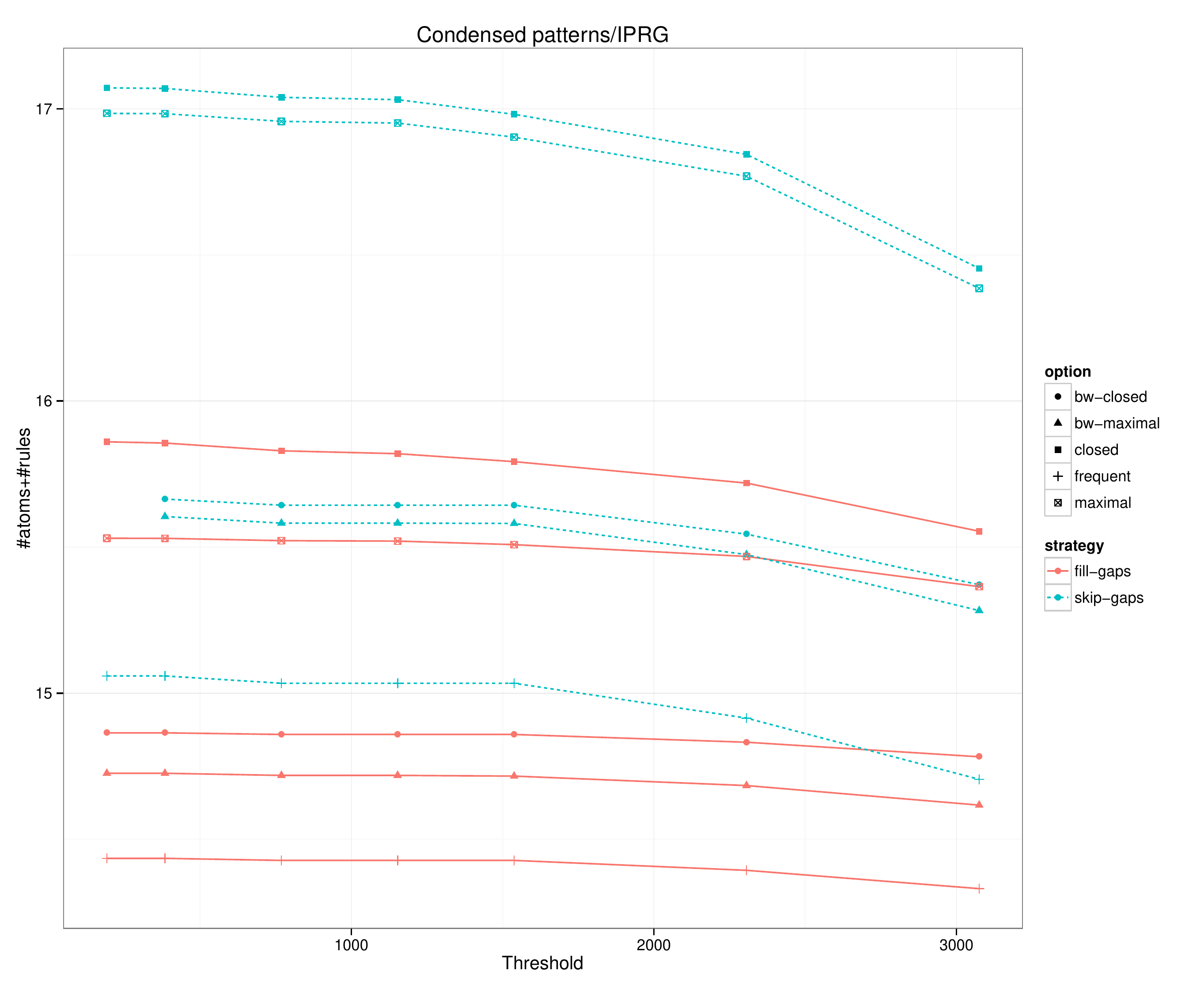}
\includegraphics[trim=0 0 109 0, clip=true, width=.32\textwidth]{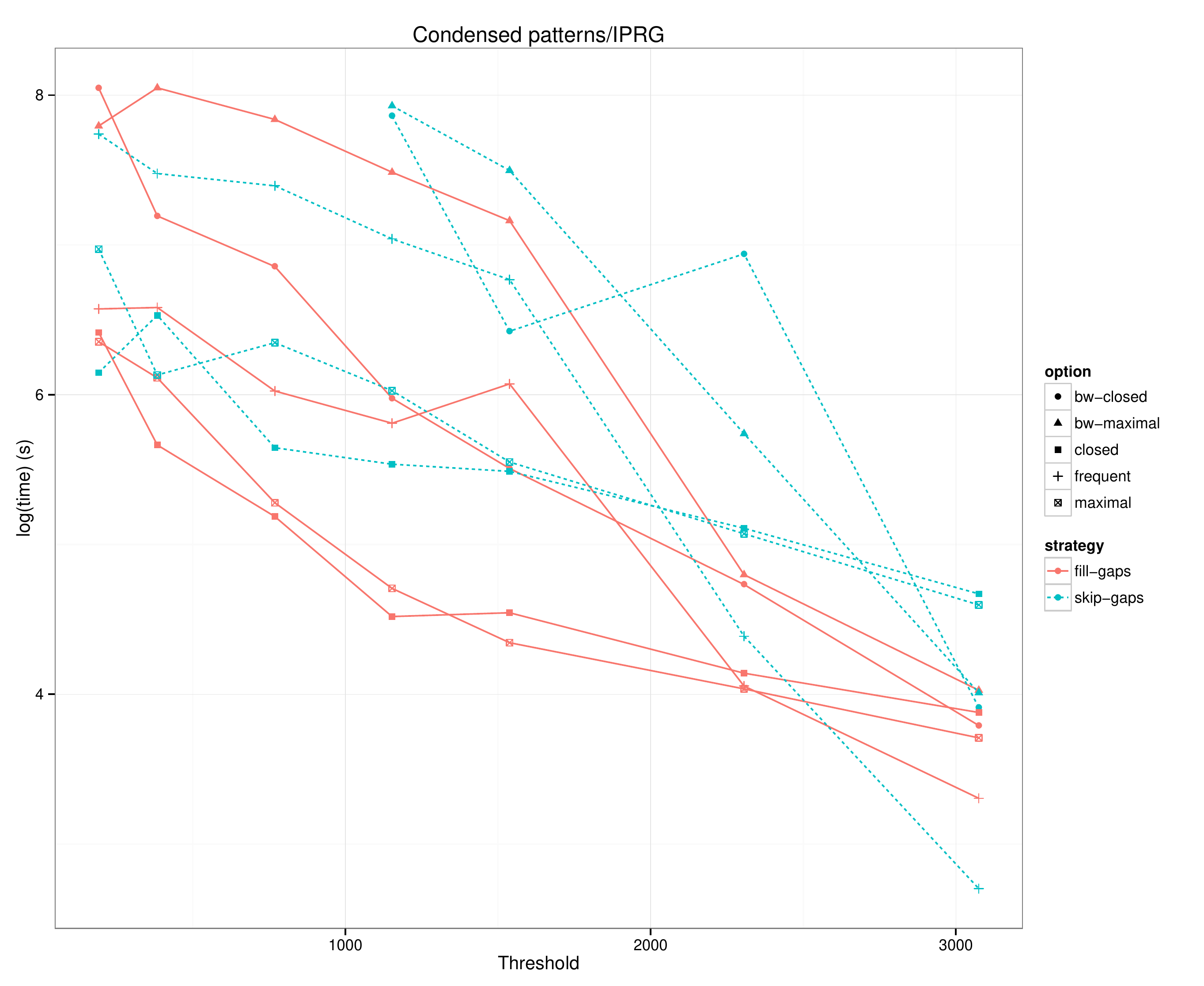}
\includegraphics[trim=0 0 109 0, clip=true, width=.32\textwidth]{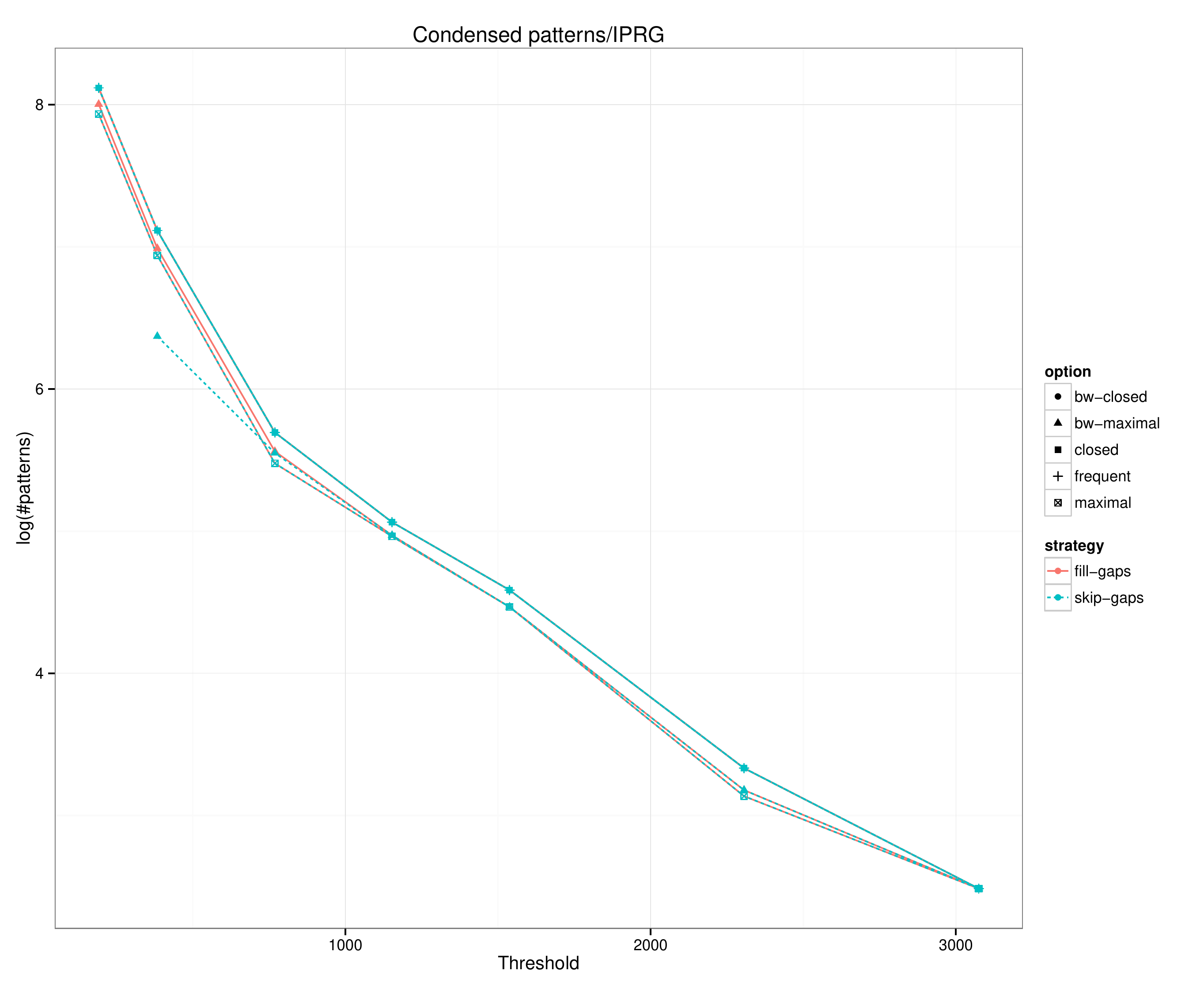}\\
\includegraphics[trim=0 0 109 0, clip=true, width=.32\textwidth]{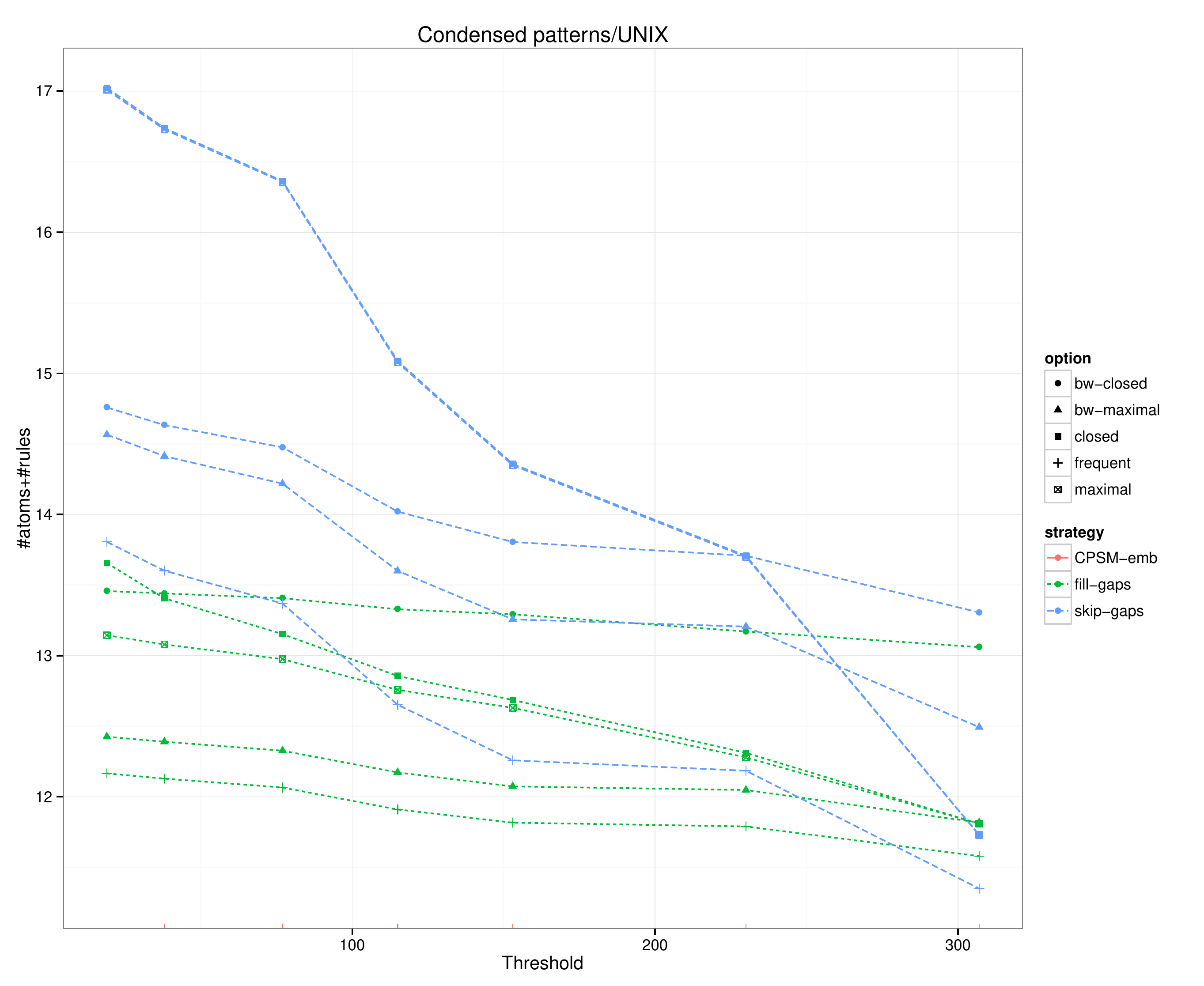}
\includegraphics[trim=0 0 109 0, clip=true, width=.32\textwidth]{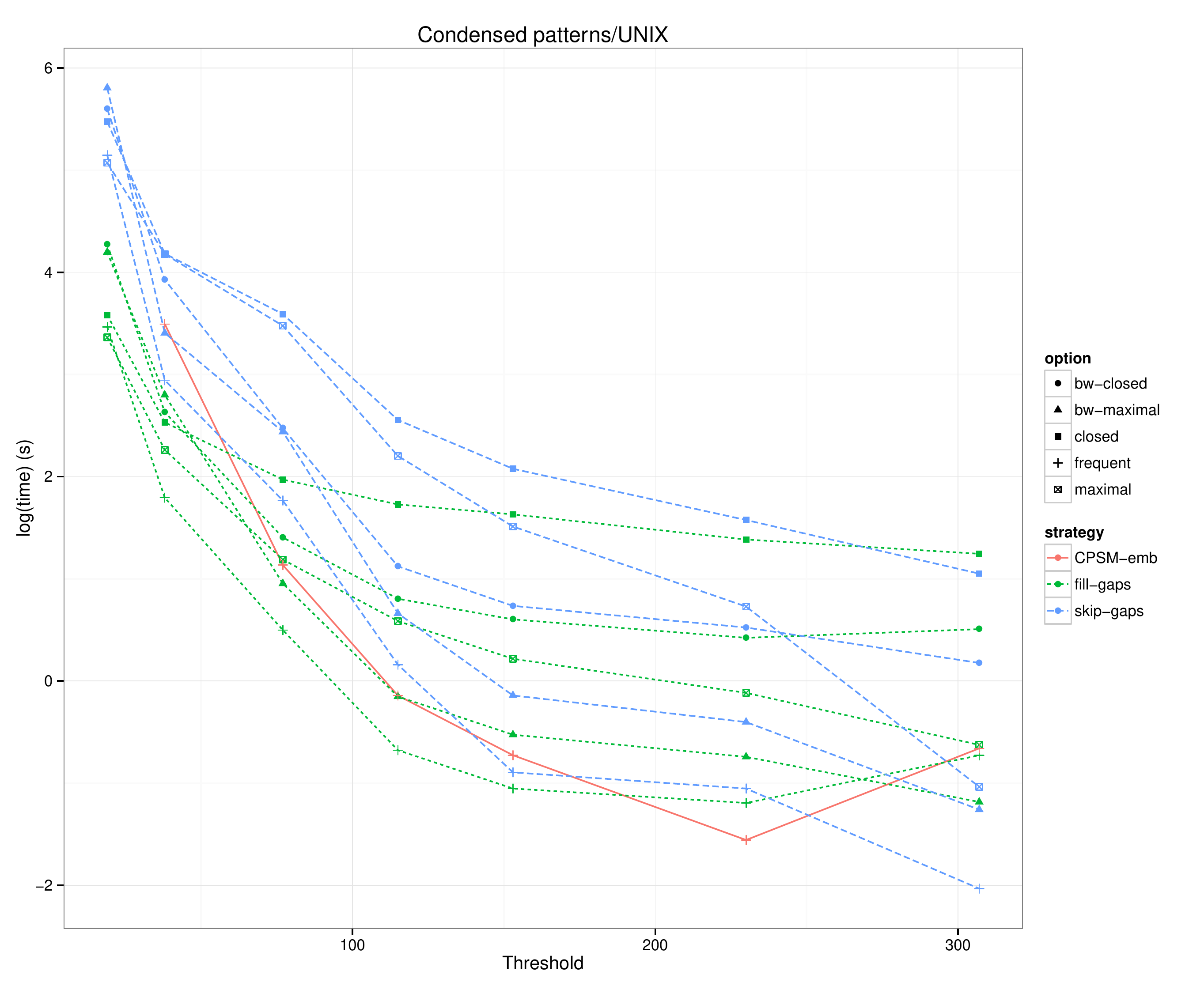}
\includegraphics[trim=0 0 109 0, clip=true, width=.32\textwidth]{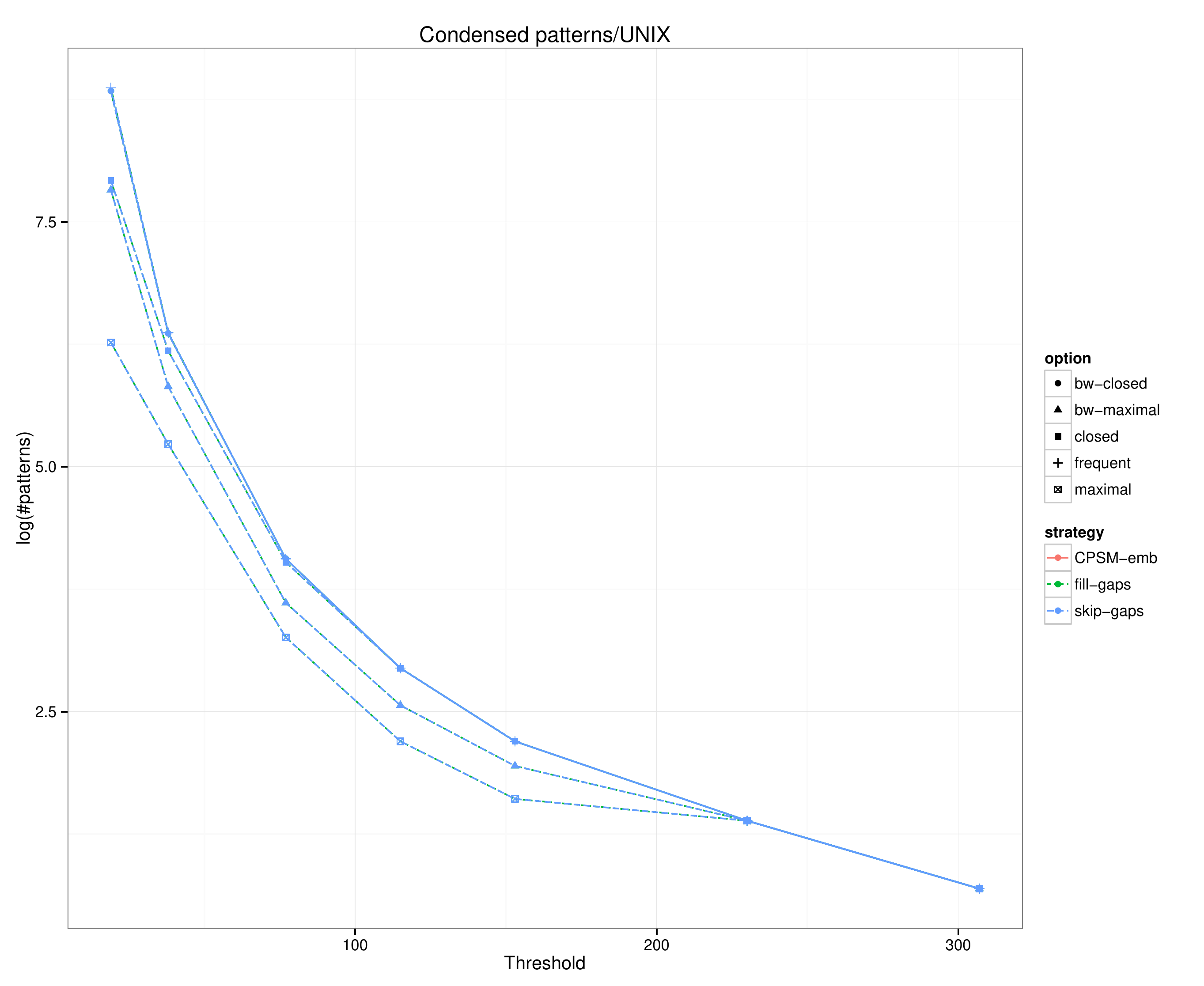}\\
\includegraphics[trim=5 58 5 58, clip=true, width=.5\textwidth]{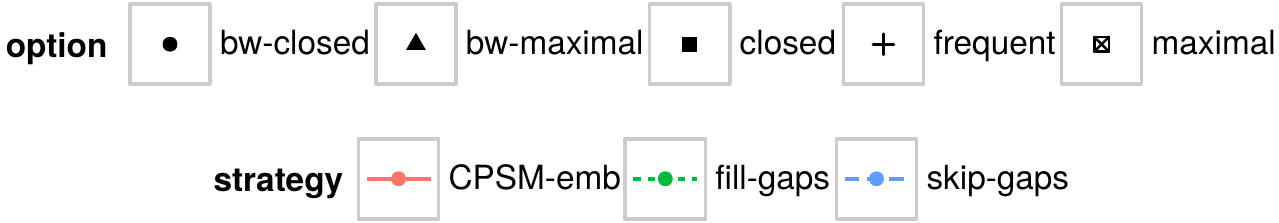}
\caption{From left to right, problem size, runtime and number of extracted patterns with respect to the frequency threshold. Runtimes are shown only if the solving was complete, contrary to pattern numbers which show the number of extracted patterns within the timeout period.
From top to bottom, JMLR, IPRG and UNIX.
For each plot, the curves illustrate the results for different types of condensed patterns (see legend) and for the two embedding strategies (\emph{fill-gaps} in red-plain line, \emph{skip-gaps} in blue-dashed line). }
\label{fig:results_condensed}
\end{figure}

Fig. \ref{fig:results_condensed} illustrates the results for condensed pattern mining. This approach cannot be compared to \cpsm\  since it does not propose means for encoding such kind of patterns.

This experiment compares the resource requirements in time and memory for mining closed/maximal and backward-closed/maximal patterns. For each of these mining task, we compared the skip-gaps and fill-gaps strategies. The main encoding is still based on prefix-projection.
Three real datasets have been processed (JMLR, UNIX and IPRG). The FIFA dataset was not processed due to its heavy memory requirement for some of these tasks.

We can first note that the difference between the number of extracted patterns is low. As expected, all encodings that complete a given mining task extract the same number of patterns. This result supports the correctness of our approach.
From the memory point of view, we see that the encodings extracting condensed patterns requires several order of magnitude more memory, especially for (backward-)closed patterns. It is also interesting to note that the memory requirement for the fill-gaps strategy is not linked to the threshold, contrary to the skip-gaps strategy. Again, the fill-gaps strategy seems to be more convenient for small thresholds. 
We can note that there is a big difference between datasets concerning runtime. For instance, frequent patterns are faster to extract for JMLR and UNIX, but maximal patterns are faster to compute on IPRG. The density of this last dataset makes maximal pattern extraction easier. Uniformly, we can conclude that fill-gaps is faster than skip-gaps. The complexity of the encoding related to insertable items with skip-gaps makes the problem difficult to solve.
Opposed to the experiments presented in \cite{Guyet_EGC2016}, we did not use any solving heuristic. For maximal patterns, a huge improvement of runtime was observed when using the \emph{subset-minimal} heuristic\footnote{The use of subset-minimal heuristic keeps solving the maximal patterns problem complete.}.

\section{Conclusion and perspectives}
This article has presented a declarative approach of sequential pattern mining based on answer set programming. We have illustrated how to encode a broad range of mining tasks (including condensed representations and constrained patterns) in pure ASP.
Thus, we shown the first advantage of declarative pattern mining: for most well-specified tasks, the development effort is significantly lower than for procedural approaches. The integration of new constraints within our framework requires only few lines of code. This was made possible thanks to the flexibility of both ASP language and solvers. 

Nonetheless, another objective of this paper was to give the intuition to the reader that while encoding a straightforward solution to a problem can be easy in ASP, writing efficient programs may be complex. Developing competitive encodings requires a good understanding of the solving process. 
To this end, we have presented several possible improvements of basic sequential pattern mining and two alternatives for encoding the main complex task, \ie computing embeddings. These encodings have been extensively evaluated on synthetic and real datasets to draw conclusions about the overall efficiency of this approach (especially compared to the constraint programming approach CPSM) and about which are the best encodings among the proposed ones and in which context.

The first conclusion of these experiments is that our ASP approach has comparable computing performances with \cpsmemb\ as long as the length of the sequence remains reasonable. This can be explained considering that solving the embedding problem is a difficult task for pure declarative encodings while \cpsm\ relies on dedicated propagators. The propagators of \cpsm\ solve the embedding problem using additional procedural code.
It turns that, for solving the embedding problem in ASP, encoding using a \emph{fill-gaps} strategy appears to be better than using the \emph{skip-gaps} strategy on real datasets, thanks to lower memory requirements.

The second conclusion is that adding constraints on patterns reduces runtime, but increases memory consumption. For real datasets, the more constraints are added, the more memory is required. This is due, to encoding the constraints, but also to encoding the information that may be required to compute constraints. For example, encodings using the \emph{maxspan} constraint require more complex embeddings (\lstinline!occS/4! atoms) than encodings without this constraint.

To fully benefit from the flexibility of our approach to proceed large datasets, we need to improve the efficiency of the computation of embeddings. Our objective is now to mimic the approach of CPSM consisting in using propagators within the solver to solve the part of the problems for which procedural approaches are efficient. The new \clingo\ 5 series will integrate ``ASP modulo theory" solving processes. This new facilities will enable to combine ASP and propagators in an efficient way.

\begin{acknowledgement}%
We would like to thanks Roland Kaminski and Max Ostrowski for their significant inputs and comments about ASP encodings; and Benjamin Negrevergne and Tias Guns for their suggestions about the experimental part. We also thank the anonymous reviewers for their valuable comments and constructive suggestions.
\end{acknowledgement}

\section*{Appendix}
\addcontentsline{toc}{section}{Appendix}
\label{sec:appendix}

 Listing \ref{list:sequences_IS} illustrates how the encoding of the \emph{skip-gaps} strategy can be transformed to mine sequential patterns that are sequences of itemsets.

\begin{lstlisting}[caption={Mining sequences with patterns as sequences of itemsets}, label=list:sequences_IS]
item(I) :- seq(_, _,I). %set of items

%sequential pattern generation
patpos(1).
0 { patpos(X+1) } 1 :- patpos(X), X<maxlen.
patlen(L) :- patpos(L), not patpos(L+1).

%generate an itemset for each position
1 { pat(X,I): item(I) } :- patpos(X).

%pattern occurrences
occS(T,1,P) :- seq(T,P,I):pat(1,I); seq(T,P,_).
occS(T,Pos+1,P) :- occS(T,Pos,Q), Q<P, seq(T,P,J),
                   pat(Pos+1,J), seq(T,P,I):pat(Pos+1,I).

support(T) :- occS(T, L, _), patlen(L).
:- { support(T) } th-1.
\end{lstlisting}

The first difference with the encoding of Listing \ref{list:sequences_sg} concerns the generation of patterns. The upper bound constraint of the choice rule in Line 9 has been removed, enabling the possible generation of every non-empty subset of $\mathcal{I}$.

The second difference is that the new ASP rules verify the inclusion of all  items in itemsets. Line 14, \lstinline!seq(T,P,I):pat(1,I)! indicates that for each atom \lstinline!pat(1,I)! there should exist an atom \lstinline!seq(T,P,I)! to satisfy the rule body. A similar expression is used Line 15.

\bibliographystyle{apalike}
 


\end{document}